\documentclass{article}

\usepackage{mathrsfs} % for mathscr
\usepackage{algorithmic}
\usepackage{algorithm}

\usepackage{amsmath}
\usepackage{amssymb}

\usepackage{color}
\usepackage{subfigure}
\usepackage{url}

\ifx\pdftexversion\undefined
  \usepackage[dvips]{graphics}
\else
  \usepackage[pdftex]{graphics}
\fi

\usepackage{fancybox}
\usepackage{version}

\includeversion{fullversion}
\excludeversion{shortversion}

\definecolor{bgcolor}{gray}{0.9}
\cornersize{0.1}
\newenvironment{competition}
{\bigskip\begin{Sbox}\begin{minipage}{\linewidth}\medskip}
{\end{minipage}\end{Sbox}\noindent\boxput*(-0.9,1){\makebox[0cm][l]{\colorbox{bgcolor}{\sc Restriction for the 2008 CSP solver competition}}}{\fbox{\TheSbox}}\medskip}

\newtheorem{definition}{Definition}
\newtheorem{remark}{Remark}

\newcommand{\xml}[1]{{\tt #1}}
\newcommand{\xmltag}[1]{{\tt <#1>}}
\newcommand{\xmlattr}[1]{{\tt\bf #1}}
\newcommand{\xmlval}[1]{{\tt "#1"}}

\begin{document}

\pagestyle{headings}
%In order to omit page numbers and running heads, please change this line to 
%\pagestyle{empty}
% and change the first command line too, see above.

%\mainmatter

\begin{fullversion}
\title{XML Representation of Constraint Networks\\ Format XCSP 2.1}
\end{fullversion}
\begin{shortversion}
\title{XML Representation of Constraint Networks\\ Format XCSP 2.1\\ \ \\
{\em Abridged Representation for the\\CPAI'08 CSP Competition}}
\end{shortversion}

\author{Olivier Roussel ~ Christophe Lecoutre
\vspace{0.5cm} \\
Universit\'e Lille-Nord de France, Artois\\
CRIL - CNRS UMR 8188 \\
IUT de Lens, F-62307 Lens \\
\{roussel,lecoutre\}@cril.fr}

%\author{Organising Committee of the \\Third International Competition of CSP Solvers}

\date{}

\maketitle

\begin{abstract}
  We propose a new extended format to represent
  constraint networks using XML.  This format allows us to represent
  constraints defined either in extension or in intension.  It also
  allows us to reference global constraints.  Any instance of the
  problems CSP (Constraint Satisfaction Problem), QCSP (Quantified
  CSP) and WCSP (Weighted CSP) can be represented using this format. 

\begin{shortversion}%
  In this document, we present an abridged version (of the format) which
  defines all constructs that will be used for the third international
  competition of CSP/Max-CSP/WCSP solvers which will be held during summer 2008
  (deadline: May 10, 2008). We also introduce the restrictions that will be enforced for the competition. 
   A companion document presents the full description of the XCSP 2.1 format.
\end{shortversion}%

\begin{fullversion}%
  A subset of this format will be used for the third international
  competition of CSP solvers which will be held during summer 2008
  (deadline: May 10, 2008).
\end{fullversion}%

\end{abstract}

The release of this document is January 15, 2008.  
%A final version is
%due in early January 2008.  Meanwhile, anyone is encouraged to make
%constructive remarks to improve the proposed format.

\pagebreak

\tableofcontents

\pagebreak

\section{Introduction}

The Constraint Programming (CP) community suffers from the lack of a standardized representation of problem instances.
This is the reason why we propose an XML representation of constraint networks.
The Extensible Markup Language (XML) \cite{XML} is a simple and flexible text format playing an increasingly important role in the exchange of a wide variety of data on the Web.
The objective of the XML representation is to ease the effort required to test and compare different algorithms by providing a common test-bed of constraint satisfaction instances.

One should notice that the proposed representation is low-level. % (however, a higher-level extension is planned). 
 More precisely, for
each instance, domains, variables, relations (if any), predicates (if
any) and constraints are exhaustively defined.
%The objective of introducing this format is to allow comparing different algorithms/solvers without any interaction with modelling and any possible interpretation of the instance. % (e.g., the order of the variables).
The current format should not be confused with powerful
modelling language such as the high-level proposals dedicated to
mathematical programming - e.g. AMPL
(\url{http://www.ampl.com}) and GAMS (\url{http://www.gams.com}) - or
dedicated to constraint programming\footnote{Remark that the specification
language Z (\url{http://vl.users.org}) has also been used to build nice (high-level) problem models \cite{RA_building}} - e.g. OPL \cite{VH_OPL}, EaCL \cite{TMWFB_EaCL}, NCL \cite{Z_NCL}, ESRA \cite{FPA_esra}, Zinc \cite{GMRW_zinc} and ESSENCE \cite{FGJMM_essence}.
Nevertheless, we also project to extend this format in a near future to take into account higher level constructs.

In this document, we present an extension, denoted XCSP 2.1, of the format XCSP 2.0 which was used for the 2006 CSP solver competition. %described in the document found at \url{http://www.cril.univ-artois.fr/XCSP}.
It aims at being a (hopefully good) compromise between readability, verbosity and structuration.
More precisely, our objective is that the representation be:
\begin{itemize}
\item readable: thanks to XML, we think that it is the case. If you want to modify an instance by hand, you can do it without too many difficulties whereas it would be almost impossible with a tabular format. Only a few constructions require an a-priori knowledge of the format. % (for example,  
\item concise: with the abridged version which doesn't use systematically XML tags and attributes, the proposed representation can be comparable in length to one that would be given in tabular format. This is important, for example, to represent instances involving constraints in extension. 
\item structured: because the format is based on XML, it remains rather easy to parse instances.  
\end{itemize}

Roughly speaking, we propose two variants of this format:
\begin{itemize}
\item a fully-tagged representation, 
\item an abridged representation.
\end{itemize}

The first representation (tagged notation) is a full
XML, completely structured representation which is suitable for using
generic XML tools but is more verbose and more tedious to write for a
human being. The second representation (abridged notation) is just a shorthand notation
of the first representation which is easier to read and write for a
human being, but less suitable for generic XML tools.

These two representations are equivalent and this document details the
translation of one representation to another. Automatic tools to
convert from one representation to another will be made available, and parsers
will accept the two representations. This will allow
human beings to use the shorthand notation to encode an instance while still being able to use generic XML tools.

% Some people might regret the fact that we have not adopted a
% fine-grained XML structuration (i.e. a full XML structuration).  On
% the one hand, it would go against our objective of conciseness and
% even readability (because, when a document is too much structured,
% it becomes too long and difficult for a human to read it).  On the
% other hand, it is always possible in the future to propose a
% fine-grained XML representation with a tool allowing to pass from
% the coarse-grained representation to the fine-grained one.  It would
% allow, for example, people to use XSLT.

\begin{shortversion}
  In this document, we (mainly) present the abridged representation of
  the format which will be adopted for the 2008 CSP solver
  competition.  More information about the fully-tagged representation
  is proposed in another document.  However, we also partially
  describe the fully-tagged representation as it makes it easier to
  understand some parts of our presentation.
\end{shortversion}

%\begin{remark}
%The format XCSP 2.1 described in this paper is an extension of the format XCSP 2.0.
%It means that an instance represented using the old format is compatible with the new one.
%\end{remark}

\section{Basic Components} 

This section describes how the most common kinds of information are
represented in this XML format. This section only details the general
data structures that are used in the description of instances. The way
these structures are used to represent an instance is presented in the
other sections of this document.
As mentioned in the introduction, we present two representations for each structure.
The first one is fully-tagged and the second one is abridged.
%Because the data structures described here are used very frequently in the description of an instance, we present two representations of these structures.

\subsection{Identifiers and Integers}

First, let us introduce the syntax of identifiers and integers.  An
identifier has to be associated with some XML elements, usually under
the form of an attribute called $name$.  An identifier must be a valid
identifier according to the most common rules (start with a letter or
underscore and further contain letters, digits or underscores).  More
precisely, identifiers and integers are defined (in BNF notation) as
follows:

{\footnotesize
\begin{verbatim}
   <identifier> ::= <letter> | "_" { <letter> | <digit> | "_" }
   <integer>  ::= [ "+" | "-"]  <digit> {<digit>}
   <letter> ::= "a".."z" | "A".."Z" 
   <digit> ::= "0".."9"
\end{verbatim}
}

Of course, identifiers are case-sensitive.

\subsection{Separators}

\subsubsection{Tagged notation}

For the tagged version, we do not need to use any specific separator
since the document is fully structured. 

\subsubsection{Abridged notation}

For the abridged version, we sometimes need to employ separators.
They are defined as follows:

{\footnotesize
\begin{verbatim}
   <whitespace> ::= " " | "\t" | "\n" | "\r"
   <separator> ::= <whitespace> | { <whitespace> }
\end{verbatim}
}

\subsection{Constants}

Different kinds of constant can be used in the encoding of a CSP
instance.

\subsubsection{Tagged notation}

Boolean constants are written using two special elements:
\xml{<true/>} and \xml{<false/>}.  Integer constants are written
inside a \xmltag{i} element (e.g. \xml{<i>19</i>}).  

\begin{fullversion}
Real constants are written inside a \xmltag{r} element (e.g. \xml{<r>19.5</r>}).
\end{fullversion}

\subsubsection{Abridged notation}

Wherever it is legal to have a numerical constant, its value can be
written directly without the enclosing tag (e.g. 19, 19.5).  19 is
considered as an integer constant. % while 19.0 is considered as a real constant.

To avoid introducing reserved keywords, Boolean constants
(\xml{<true/>} and \xml{<false/>}) cannot be abridged.

\subsection{Intervals}

\subsubsection{Tagged notation}

Intervals are represented by a \xmltag{interval} element with two
attributes: \xmlattr{min} and \xmlattr{max}. The \xmlattr{min}
represents the minimal value of the interval (the lower bound) and the
\xmlattr{max} attribute the maximal value (the upper bound). For
example, \xml{<interval min="10" max="13"/>} corresponds to the set
$\{10,11,12,13\}$.

\subsubsection{Abridged notation}

To represent an interval, on has just to write two constants (of the same type) separated by the sequence "..".
For example, 10..13 corresponds to the set of integers $\{10,11,12,13\}$.

\subsection{Variables}
\label{basictypes:variables}

Several constructs in the format have to reference variables.  For
simplicity, we consider here that the term variable refers to both
effective and formal parameters of functions and predicates.

\subsubsection{Tagged notation}

The reference to a variable is represented by a \xmltag{var} element
with an empty body and a single attribute \xmlattr{name} which
provides the identifier of the variable.  For example, a reference to
the variable $X1$ is represented by \xml{<var name="X1"/>}.

\subsubsection{Abridged notation}

Wherever it is legal to have a \xml{<var name="\emph{identifier}"/>} element, this element can be replaced equivalently by \xml{\emph{identifier}}. 
For example, \xml{X1} and \xml{<var name="X1"/>} are two legal and equivalent ways to refer to the variable \xml{X1}.

\subsection{Formal parameters}
\label{basictypes:formalparms}

A predicate or function in this XML format must first define the list
of its formal parameters (with their type). Then, these formal
parameters can be referenced with the notations defined in section
\ref{basictypes:variables}.

In both tagged and abridged representations, formal parameters are
defined in the body of a \xmltag{parameters} element.

\subsubsection{Tagged notation}

In the tagged notation, each formal parameter is defined by a
\xmltag{parameter} element with two attributes. The attribute
\xmlattr{name} defines the formal name of the parameter and the
attribute \xmlattr{type} defines its type. 
For example, \xml{<parameter  name="X0" type="int">} defines a parameter named \xml{X0} of integer type.

\subsubsection{Abridged notation}

Wherever it is legal to have a \xmltag{parameter} element, a formal
parameter can be written in abridged notation by its type followed by
whitespace followed by the parameter name (as in C and Java programming language). Consecutive parameters must
be separated by whitespace. For example, \xml{int X0} is the abridged
representation of \xml{<parameter name="X0" type="int">}.

The syntax of the formal parameters list is described by the following
grammar (in BNF notation):

{\footnotesize
\begin{verbatim}
   <formalParameters> ::= [<formalParametersList>]
   <formalParametersList> ::= <formalParameter>  
                            | <formalParameter> <separator> <formalParametersList>
   <formalParameter> ::= <type> <separator> <identifier>
   <type>  ::= "int"  
\end{verbatim}
}

%Note that more than any number of white spaces is allowed between formal parameters and between the type and the name of a parameter.

\subsection{Lists}
\label{basictypes:list}

A list is an array of (possibly heterogeneous) objects.  The order of
objects in the list is significant (but this order may be deliberately
ignored when needed).

\subsubsection{Tagged notation}

A list is represented by a \xmltag{list} element with all members of the list given in the body of the element. 
For example, a list containing the integers $1$, $2$ and the Boolean value $true$ is represented by:

\xml{<list> <i>1</i> <i>2</i> <true/> </list>}

\subsubsection{Abridged notation}

Wherever it is legal to have a list element, the opening square brace
is defined as a synonym of \xml{<list>} and the closing square brace
is defined as a synonym of \xml{</list>}.
A separator is used between two elements of the list.
Whitespace can be found before and after a square brace.

\xml{[1 2 <true/>]}

\begin{fullversion}
\subsection{Matrices}

Vectors are represented as lists, 2-dimensions matrices are
represented as lists of vectors, 3-dimensions matrices are represented
as lists of 2-dimensions matrices and so on. To improve readability,
2-dimensions matrices are represented as lists of lines of
coefficients.

\begin{verbatim}
[1 2 3]

[
 [1 2 3]
 [0 1 2]
 [0 0 1]
]
\end{verbatim}
\end{fullversion}

\subsection{Dictionaries}

% key is only a string

A dictionary is an associative array that maps a key to a value. In
other words, it is an array of $<$key,value$>$ pairs. A key is a name
which references a value in the data structure. A notation common to a
number of languages to access the value corresponding to a key $k$ in
a dictionary $d$ is $d[k]$. In a sense, a dictionary is a
generalization of an array: indices in (classical) arrays must be contiguous
integers while keys in a dictionary can be arbitrary names. A
dictionary can also be seen as a generalization of the notion of
structures (struct in C/C++) and records (Pascal). A record with $n$
fields $f_1,\ldots,f_n$ can be seen as a dictionary containing the
$n$ keys $f_1,\ldots,f_n$ and the corresponding values. A dictionary
is an extension of a record since new keys can be added to a
dictionary while a record usually has a fixed list of fields.
Each pair $<$key,value$>$ in a dictionary is called an entry in that
dictionary.

A function which accepts a dictionary as parameter can decide that
some keys must be present in the dictionary and that some others keys
are optional. This provides a simple way to support optional
parameters. When a key is missing in a dictionary, it is considered
that there is no corresponding value. The special tag \xml{<nil/>} is
another way to specify explicitly that a key has no corresponding
value. This special value corresponds to the {\tt null} value in SQL.
Omitting a key $k$ from a dictionary or defining that key $k$
corresponds to value \xml{<nil/>} are equivalent ways of associating
no value to key $k$.

The order of keys in a dictionary is not significant. A dictionary may
contain no key at all. A dictionary can be associated to a key in a
given dictionary (in other words, dictionaries may be contained in a
dictionary).

\subsubsection{Tagged notation}

An entry of a dictionary that associates a value $v$ with key $k$ is
encoded by an element \xmltag{entry} with a single attribute
\xmlattr{key} with value $k$ and a body containing the value $v$:
\xml{<entry key="k">v</entry>}

A dictionary is defined by a \xmltag{dict} element with all entries of
the dictionary given inside the body: \xml{<dict><entry
  key="name1">value1</entry> <entry
  key="name2">value2</entry></dict>}. The body of this element cannot contain other elements than dictionary entries.

\subsubsection{Abridged notation}

Several notations are already used in different languages to associate
a key with a value in an associative array ({\tt key => value} in PHP
and Perl, {\tt /key value} in PostScript and PDF,...). Since the
character $>$ is a reserved character in XML, we use the PostScript
notation.

A dictionary in abridged notation starts by a opening curly brace
followed by a list of entries and is ended by a closing curly
brace. Each entry is written as a key immediately preceded by a slash
(no space between the slash and the key), whitespace and the value
corresponding to this key.  Whitespace can be found before and after a
curly brace. For example:

\xml{\{/name1 value2 /name2 value2\}}

\subsubsection{Conventional order}

In some cases (e.g. when a function expects a dictionary with a given
set of keys), a conventional order can be associated with a
dictionary. This conventional order specifies a default order of keys
which can be used to further shorten the notations. When the
conventional order of keys can be known from the context, a dictionary
can be written in abridged notation by opening a curly brace, listing
the values of each key which is expected in the dictionary and closing
the curly brace. The absence of a slash following the opening curly
brace identifies a dictionary represented in conventional order (note
however that white space is allowed between the opening curly brace
and the first slash).

In this context, there must be as many values inside the curly braces
as the number of keys in the conventional order. To assign no value to
a given key, the special value \xml{<nil/>} must be used.

For example, the coordinates of a point of a plane may be represented
by a dictionary containing two keys $x$ and $y$. The point at
coordinates $(2,5)$ can be represented by several notations. 
We can have:

\begin{verbatim}
<dict>
  <entry key="x"><i>2</i></entry>
  <entry key="y"><i>5</i></entry>
</dict>

or:

<dict>
  <entry key="y"><i>5</i></entry>
  <entry key="x"><i>2</i></entry>
</dict>

or:

{/x 2 /y 5}

or:

{/y 5 /x 2}

or:

{
  /x 2
  /y 5
}
\end{verbatim}

When a conventional order is fixed which indicates that key $x$ is given before key $y$, the same dictionary can be written by:

\begin{verbatim}
{2  5}
\end{verbatim}

\subsection{Tuples} \label{sec:tuples}

Here, we consider a tuple as being a sequence of objects of the same type.
For example, $(2,5,8)$ is a tuple containing three integers.

\subsubsection{Tagged notation}

A tuple is represented by a \xmltag{tuple} element with all members of the tuple given in the body of the element.
For example, the tuple  $(2,5,8)$ is represented by:

\xml{<tuple> <i>2</i> <i>5</i> <i>8</i> </tuple>}

\subsubsection{Abridged notation}

For the abridged variant, the members of any tuple are written directly within the enclosing tag.
However, if we have two successive tuples (i.e. \xml{... </tuple> <tuple> ...}), we use the character '$|$' as a separator between them.
For example, the representation of a sequence of binary tuples takes the form (in BNF notation):

{\footnotesize
\begin{verbatim}
   <binaryTupleSequence> ::= <binaryTuple> | <binaryTuple> "|" <binaryTupleSequence>
   <binaryTuple> ::= <integer> <separator> <integer>
\end{verbatim}
}

For ternary relations, one has just to consider tuples formed from $3$ values, etc. 
For example, a list of binary tuples is:

{\footnotesize
\begin{verbatim}
    0 1|0 3|1 2|1 3|2 0|2 1|3 1
\end{verbatim}
}
while a list of ternary tuples is:
{\footnotesize
\begin{verbatim}
    0 0 1|0 2 1|1 0 1|1 2 0|2 1 1|2 2 2
\end{verbatim}
}

\subsection{Weighted Tuples} \label{sec:wtuples}

It may be interesting (e.g. see the WCSP framework \cite{L_node}) to associate a weight (or cost) with a tuple or a sequence of tuples.

\subsubsection{Tagged notation}

We then just have to enclose this tuple (these tuples) within a \xmltag{weight} element which admits one attribute \xmlattr{value}. 
This attribute must be an integer or the special value "infinity".
For example, if a weight equal to $10$ must be associated with the tuple  $(2,5,8)$, we obtain:

\xml{<weight value="10"> <tuple> <i>2</i> <i>5</i> <i>8</i> </tuple> </weight>}

Notice that it is possible to directly associate the same weight with several tuples.

\subsubsection{Abridged notation}

In abridged notation, each tuple can be given an explicit cost by prefixing it with its cost followed by a colon character ':'.  
When a cost is not specified for a tuple, it simply means that the cost of the current tuple is equal to the cost of the previous one.  
At the extreme, only the first tuple is given an explicit cost, all other tuples implicitly referring to this cost.
In any case, the first tuple of a relation must be given an explicit cost.
Remark that with the abridged variant, it is not possible to put in the same context unweighted and weighted tuples (but, we believe that it is not a real problem).
Finally, to associate the special value "infinity" with a tuple, the special element \xml{<infinity/>} must be used.

\smallskip

For example, let us consider the following ``classical'' list of binary tuples:

{\footnotesize
\begin{verbatim}
    0 1|0 3|1 2|1 3|2 0|2 1|3 1
\end{verbatim}
}

If $1$ is the cost of tuples $(0, 1), (0, 3), (3, 1)$ whereas $10$ is the cost of all other tuples, then we can write:

{\footnotesize
\begin{verbatim}
    1:0 1|1:0 3|10:1 2|10:1 3|10:2 0|10:2 1|1:3 1
\end{verbatim}
} 

but also, using implicit costs:

{\footnotesize
\begin{verbatim}
    1:0 1|0 3|10:1 2|1 3|2 0|2 1|1:3 1
\end{verbatim}
} 

This example may also be written equivalently on several lines:
{\footnotesize
\begin{verbatim}
    1:  0 1|0 3|
    10: 1 2|1 3|2 0|2 1|
    1:  3 1
\end{verbatim}
} 

Note that using the abridged representation to associate costs with tuples allows us to save a large amount of space.  

%\begin{fullversion}
%\subsection{Summary of special characters}
%\end{fullversion}

\section{Representing CSP instances}

%For the second international competition of CSP solvers which will be held during summer 2006, it has been decided to deal with constraint networks involving finite domains and constraints defined in extension or in intension.
%It means that, in such networks, domains (associated with variables) correspond to finite sets of values, and constraints are either explicitly defined by sets of tuples, or implicitly defined by predicates.

In order to avoid any ambiguity, we briefly introduce constraint networks.
A constraint network consists of a finite set of variables such that each variable $X$ has an associated domain $dom(X)$ denoting the set of values allowed for $X$, and a finite set of constraints such that each constraint $C$ has an associated relation $rel(C)$ denoting the set of tuples allowed for the variables $scp(C)$ involved in $C$. 
A solution to a constraint network is the assignment of a value to each variable such that all the constraints are satisfied. 
A constraint network is said to be satisfiable if it admits at least a solution.
The Constraint Satisfaction Problem (CSP), whose task is to determine whether or not a given constraint network is satisfiable, is NP-hard.
A constraint network is also called a CSP instance.
For an introduction to constraint programming, see for example \cite{D_book,A_principles}.

Each CSP instance is represented following the format given in Figure \ref{fig:xml} where $q$, $n$, $r$, $p$ and $e$ respectively denote the number of distinct domains, the number of variables, the number of distinct relations, the number of distinct predicates and the number of constraints.
Note that $q \leq n$ as the same domain definition can be used for different variables, $r \leq e$ and $p \leq e$ as the same relation or predicate definition can be used for different constraints.
Thus, each instance is defined by an XML element which is called \xmltag{instance} and which contains four, five or six elements.
Indeed, it is possible to have one instance defined without any reference to a relation or/and to a predicate.
Then, the elements \xmltag{relations} and \xmltag{predicates} may be missing (if both are missing, it means that only global constraints are referenced). 

\begin{figure}[p]
{\footnotesize
\begin{verbatim}
<instance>
   <presentation
       name = 'put here the instance name'
       ...
       format = 'XCSP 2.1' > 
     Put here the description of the instance
   </presentation>

   <domains nbDomains='q'>
      <domain 
         name = 'put here the domain name'
         nbValues = 'put here the number of values' >
        Put here the list of values
      </domain>
      ...
   </domains>

   <variables nbVariables='n'>
      <variable 
         name = 'put here the variable name'
         domain = 'put here the name of a domain' 
      />
      ...
   </variables> 

   <relations nbRelations='r'>
       <relation 
           name = 'put here the name of the relation'
           arity = 'put here the arity of the relation'
           nbTuples = 'put here the number of tuples'
           semantics = 'put here either supports or conflicts' > 
         Put here the list of tuples
       </relation> 
        ...
   </relations>

   <predicates nbPredicates='p'>
       <predicate 
           name = 'put here the name of the predicate' >
         <parameters>
           put here a list of formal parameters
         </parameters>
         <expression>
            Put here one (or more) representation of the predicate expression
         </expression> 
       </predicate>
       ...
   </predicates>

   <constraints nbConstraints='e'>
      <constraint 
         name = 'put here the name of the constraint'
         arity = 'put here the arity of the constraint' 
         scope = 'put here the scope of the constraint' 
         reference = 'put here the name of a relation, a 
                      predicate or a global constraint'>
         ...
      </constraint>
      ...
   </constraints>
</instance>
\end{verbatim}
\caption{XML representation of a CSP instance \label{fig:xml}}
}
\end{figure}

Each basic element (\xmltag{presentation}, \xmltag{domain}, \xmltag{variable}, \xmltag{relation}, \xmltag{predicate} and \xmltag{constraint}) of the representation admits an attribute called \xmlattr{name}.
The value of the attribute \xmlattr{name} must be a valid identifier (as introduced in the previous section). 
In the representation of any instance, it is not possible to find several attributes "name" using the same identifier.

\begin{remark}
In the body of any element of the document, one can insert an \xmltag{extension} element in order to put any information specific to a solver.
\end{remark}

\subsection{Presentation}

The XML element called \xmltag{presentation} admits a set of attributes and may contain a description (a string) of the instance:

{\footnotesize
\begin{verbatim}
   <presentation
       name = 'put here the instance name'
       maxConstraintArity = 'put here the greatest constraint arity' 
       minViolatedConstraints = 'the minimum number of violated constraints'
       nbSolutions = 'put here the number of solutions'     
       solution = 'put here a solution'
       type = 'CSP'>
       format = 'XCSP 2.1' 
     Put here the description of the instance
   </presentation>
\end{verbatim}
}

Only the attribute \xmlattr{format} is mandatory (all other attributes are optional as they mainly provide human-readable information).
It must be given the value 'XCSP 2.1' for the current format.
The attribute \xmlattr{name} must be a valid identifier (or the special value '?').
The attribute \xmlattr{maxConstraintArity} is of type integer and denotes the greatest arity of all constraints involved in the instance.
The attribute \xmlattr{minViolatedConstraints} can be given an integer value denoting the minimum number of constraints that are violated by any full instantiation of the variables, an expression of the form 'at most k' with k being a positive integer or '?'.
The attribute \xmlattr{nbSolutions} can be given an integer value denoting the total number of solutions of the instance, an expression of the form 'at least k' with k being a positive integer or '?'.

For example,  
\begin{itemize}
\item \verb|nbSolutions = '0'| indicates that the instance is unsatisfiable,
\item \verb|nbSolutions = '3'| indicates that the instance has exactly $3$ solutions,
\item \verb|nbSolutions = 'at least 1'| indicates that the instance has at least $1$ solution (and, hence, is satisfiable),
\item \verb|nbSolutions = '?'| indicates that it is unknown whether or not the instance is satisfiable,
\end{itemize}

The attribute \xmlattr{solution} indicates a solution if one exists
and has been found.  The \xmlattr{type} attribute indicates the kind
of problem described by this instance.  It should be set to 'CSP' for
a constraint satisfaction problem, 'QCSP' for a quantified CSP and
'WCSP' for a weighted CSP.  For compatibility with the XCSP 2.0 format,
this attribute is optional for CSP instances (but it is strongly
advised to use it).  It is mandatory for other types of instances
(QCSP, WCSP,...).

\begin{shortversion}
\begin{competition}
  Only the attributes \xmlattr{maxConstraintArity} and
  \xmlattr{format} of element \xmltag{presentation} will be present
  (and the content of element \xmltag{presentation} empty).
  Potentially, the attribute \xmlattr{type} will be present.
\end{competition}
\end{shortversion}

Remark that the optional attribute
\xmlattr{maxSatisfiableConstraints}, although still authorized, is
deprecated.  We encourage to use \xmlattr{minViolatedConstraints}
instead.

\subsection{Domains}

The XML element called \xmltag{domains} admits an attribute which is
called \xmlattr{nbDomains} and contains some occurrences (at least,
one) of an element called \xmltag{domain}, each one being associated
with at least one variable of the instance.  The attribute
\xmlattr{nbDomains} is of type integer and its value is equal to the
number of occurrences of the element \xmltag{domain}.  Each element
\xmltag{domain} admits two attributes, called \xmlattr{name} and
\xmlattr{nbValues} and contains a list of values, as follows:

{\footnotesize
\begin{verbatim}
   <domain 
       name = 'put here the domain name'
       nbValues = 'put here the number of values' >
     Put here the list of values
   </domain>
\end{verbatim}
}

The attribute \xmlattr{name} corresponds to the name of the domain and its value must be a valid identifier. 

\begin{shortversion}
\begin{competition}
The value of the attribute \xmlattr{name} of the  $i^{th}$ element \xmltag{domain} must be the letter $D$ followed by the  number $i-1$ (i.e. we have $D0$, $D1$, ...).
Besides, each domain must be referenced by at least one variable.
\end{competition}
\end{shortversion}

The attribute \xmlattr{nbValues} is of type integer and its value is equal to
the number of values of the domain.  The content of the element
\xmltag{domain} gives the set of integer values included in the domain.
More precisely, it contains a sequence of integers and integer
intervals.

For the abridged variant, we have for example: %, in abridged notation:
\begin{itemize}
\item 1 5 10 corresponds to the set $\{1,5,10\}$.
\item 1..3 7 10..14 corresponds to the set $\{1,2,3,7,10,11,12,13,14\}$.
\end{itemize}

Note that \xmlattr{nbValues} gives the number of values of the domain (i.e. the domain size), and not, the number of domain pieces (integers and integer intervals).
%Note also that one (or more) space character is used as a separator of domain pieces. 

\begin{shortversion}
\begin{competition}
The values that an element \xmltag{domain} contains must be given in ascending order without multiple occurrences of the same value.
\end{competition}
\end{shortversion}

\subsection{Variables}

The XML element called \xmltag{variables} admits an attribute which is called \xmltag{nbVariables} and contains some occurrences (at least, one) of an element called \xmltag{variable}, one for each variable of the instance.
The attribute \xmlattr{nbVariables} is of type integer and its value is equal to the number of occurrences of the element \xmltag{variable}. 
Each element \xmltag{variable} is empty but admits two attributes, called  \xmlattr{name} and  \xmlattr{domain}, as follows:

{\footnotesize
\begin{verbatim}
   <variable 
      name = 'put here the variable name'
      domain = 'put here the name of a domain'   
   />
\end{verbatim}
}

The attribute  \xmlattr{name} corresponds to the name of the variable and its value must be a valid identifier. 

\begin{shortversion}
\begin{competition}
The value of the attribute  \xmlattr{name} of the  $i^{th}$ element \xmlattr{variable} must be the letter $V$ followed by the number $i-1$ (i.e. we have $V0$, $V1$, ...).
Besides, each variable must be referenced by at least one constraint (i.e. must be involved in at least one constraint).
\end{competition}
\end{shortversion}

The value of the attribute  \xmlattr{domain} gives the name of the associated domain.
It must correspond to the value of the  \xmlattr{name} attribute of a \xmlattr{domain} element. 

\subsection{Relations}

If present, the XML element called \xmltag{relations} admits an attribute which is called \xmlattr{nbRelations} and contains some occurrences (at least, one) of an element called \xmltag{relation}, each one being associated with at least one constraint of the instance.
The attribute \xmlattr{nbRelations} is of type integer and its value is equal to the number of occurrences of the element \xmltag{relation}. 

Each element \xmltag{relation} admits four attributes, called \xmlattr{name},
\xmlattr{arity}, \xmlattr{nbTuples} and \xmlattr{semantics}, and contains a list of distinct
tuples that represents either allowed tuples (supports) or disallowed
tuples (conflicts).  It is defined as follows:

{\footnotesize
\begin{verbatim}
   <relation 
       name = 'put here the name of the relation'
       arity = 'put here the arity of the relation'
       nbTuples = 'put here the number of tuples'
       semantics = 'put here either supports or conflicts' > 
     Put here the list of tuples
   </relation> 
\end{verbatim}
}

The attribute \xmlattr{name} corresponds to the name of the relation and its value must be a valid identifier. 

\begin{shortversion}
\begin{competition}
The value of the attribute \xmlattr{name} of the  $i^{th}$ element \xmltag{relation} must be the letter $R$ followed by the number $i-1$ (i.e. we have $R0$, $R1$, ...).
Besides, each relation must be referenced by at least one constraint.
\end{competition}
\end{shortversion}

The attribute \xmlattr{arity} is of type integer and its value is equal to the
arity of the relation.  The attribute \xmlattr{nbTuples} is of type integer
and its value is equal to the number of tuples of the relation.  The
attribute \xmlattr{semantics} can only be given two values: 'supports' and
'conflicts'.  Of course, if the value of \xmlattr{semantics} is 'supports'
(resp. 'conflicts'), then it means that the list of tuples correspond
to allowed (resp. disallowed) tuples.  The content of the element
\xmltag{relation} gives the set of distinct tuples of the relation.

The representation of lists of tuples is given in Section \ref{sec:tuples}.

\begin{fullversion}
  Note that an empty list of tuples is authorized by the syntax: a
  relation with an empty list of tuples is trivially unsatisfiable if
  its semantics is 'supports', and trivially satisfied if its
  semantics is 'conflicts'.
\end{fullversion}

\begin{shortversion}
\begin{competition}
  It is imposed that the tuples contained in an element \xmltag{relation}
  must be given in ascending (lexicographic) order without multiple occurrences of the same tuple.
  Also, a relation must contain at least one tuple (otherwise the relation
  is either trivially unsatisfiable or trivially satisfied, depending
  on its semantics).
\end{competition}
\end{shortversion}

%\ifthenelse{\boolean{full}}{
%\begin{remark}
%Strictly speaking, an element $<$relation$>$ does not correspond to a well-defined relation.
%Indeed, the Cartesian product on which the relation should be defined is not precised.
%However, it allows to associate an element $<$relation$>$ with constraints whose domains (i.e. Cartesian products corresponding to their scopes) are different.
%We could even have removed the attribute $<$semantics$>$, but it would have not render the , and it is very unlikely that a list of supports for one constraint corresponds to a list of conflicts for another contraint.
%\end{remark}
%}
 
\subsection{Predicates} \label{sec:predicates}

If present, the XML element called \xmltag{predicates} admits an attribute which is called \xmlattr{nbPredicates} and contains some occurrences (at least, one) of an element called \xmltag{predicate}, one for each predicate associated with at least a constraint of the instance.
The attribute \xmlattr{nbPredicates} is of type integer and its value is equal to the number of occurrences of the element \xmltag{predicate}. 

Each element \xmltag{predicate} admits one attribute, called \xmlattr{name}, and contains two elements, called \xmltag{parameters} and \xmltag{expression}.
It is defined as follows:

{\footnotesize
\begin{verbatim}
   <predicate 
       name = 'put here the name of the predicate' >
     <parameters>
        put here a list of formal parameters
     </parameters>
     <expression>
        Put here one (or several) representation(s) of the predicate expression
     </expression> 
   </predicate>
\end{verbatim}
}

The attribute \xmlattr{name} corresponds to the name of the predicate and its value must be a valid identifier. 

\begin{shortversion}
\begin{competition}
The value of the attribute \xmlattr{name} of the $i^{th}$ element \xmltag{predicate} must be the letter $P$ followed by the number $i-1$ (i.e. we have $P0$, $P1$, ...).
Besides, each predicate must be referenced by at least one constraint.
\end{competition}
\end{shortversion}

The \xmltag{parameters} element defines the list of formal parameters of the predicate. The syntax of this element is detailed is section \ref{basictypes:formalparms}.

\begin{shortversion}
\begin{competition}
The name of the $i^{th}$ formal parameter must be the letter $X$ followed by the number $i-1$ (i.e. we have  $X0$, $X1$, ...).
\end{competition}
\end{shortversion}

The only authorized type is for the moment 'int' (denoting integer values).

\begin{fullversion}
However, in the future, other types will be taken into account:  "bool", "string", etc.
\end{fullversion}

The element \xmltag{expression} may contain several representations of
the predicate expression.

\begin{shortversion}
\begin{competition}
  Only the functional representation, described below, is
  considered. Each formal parameter occurs at least one time in the
  expression of the predicate.
\end{competition}
\end{shortversion}

\subsubsection{Functional Representation}

It is possible to insert a functional representation of the predicate expression by inserting in \xmltag{expression} an element \xmltag{functional} which contains any Boolean expression defined as follows:
                  
{\footnotesize
\begin{verbatim}
<integerExpression> ::= 
      <integer> | <identifier>
    | "neg(" <integerExpression> ")"  
    | "abs(" <integerExpression> ")" 
    | "add(" <integerExpression> "," <integerExpression> ")" 
    | "sub(" <integerExpression> "," <integerExpression> ")" 
    | "mul(" <integerExpression> "," <integerExpression> ")" 
    | "div(" <integerExpression> "," <integerExpression> ")" 
    | "mod(" <integerExpression> "," <integerExpression> ")" 
    | "pow(" <integerExpression> "," <integerExpression> ")" 
    | "min(" <integerExpression> "," <integerExpression> ")" 
    | "max(" <integerExpression> "," <integerExpression> ")" 
    | "if(" <booleanExpression> "," <integerExpression>  "," <integerExpression> ")"
\end{verbatim}

\begin{verbatim}
<booleanExpression> ::= 
      "false" | "true" 
    | "not(" <booleanExpression> ")" 
    | "and(" <booleanExpression> "," <booleanExpression> ")" 
    | "or(" <booleanExpression> "," <booleanExpression> ")" 
    | "xor(" <booleanExpression> "," <booleanExpression> ")"
    | "iff(" <booleanExpression> "," <booleanExpression> ")"
    | "eq(" <integerExpression> "," <integerExpression> ")" 
    | "ne(" <integerExpression> "," <integerExpression> ")" 
    | "ge(" <integerExpression> "," <integerExpression> ")" 
    | "gt(" <integerExpression> "," <integerExpression> ")" 
    | "le(" <integerExpression> "," <integerExpression> ")" 
    | "lt(" <integerExpression> "," <integerExpression> ")" 
\end{verbatim}
}

Hence, any constraint in intension can be defined by a predicate which corresponds to an expression built from (Boolean and integer) constants and the introduced set of functions (operators).
The semantics of operators is given by Table \ref{tab:semantics}.
%Overflows, divisions by zero and integer divisions are discussed in the document dealing with the format rules for the competition.

\begin{table}[hbt]
\begin{center}
{\footnotesize
\begin{tabular}{|c|c|c|c|c|} \hline 
Operation & Arity & Syntax & Semantics & MathML \\
\hline 
\multicolumn{2}{c}{ } \\
\multicolumn{5}{l}{Arithmetic (operands are integers)} \\
\hline 
Opposite  & 1 & neg(x) & -x & $<$minus$>$ \\
Absolute Value &  1 & abs(x) & $\mid$ x $\mid$ & $<$abs$>$\\
Addition     & 2 & add(x,y) & x + y & $<$plus$>$ \\
Substraction & 2 & sub(x,y) & x - y & $<$minus$>$ \\
multiplication  & 2 & mul(x,y) & x * y & $<$times$>$ \\
Integer Division & 2 & div(x,y) & x div y & $<$quotient$>$ \\
Remainder  & 2 & mod(x,y) & x mod y & $<$rem$>$ \\ 
Power  & 2 & pow(x,y) & x$^{y}$ & $<$power$>$ \\
Minimum  & 2 & min(x,y) & min(x,y) & $<$min$>$ \\
Maximum  & 2 & max(x,y) & max(x,y) & $<$max$>$ \\
\hline      
\multicolumn{2}{c}{ } \\
\multicolumn{5}{l}{Relational (operands are integers)} \\
\hline 
Equal to     & 2 & eq(x,y) & x = y & $<$eq$>$ \\
Different from    & 2 &  ne(x,y) & x $\neq$ y & $<$neq$>$ \\
Greater than or equal     & 2 & ge(x,y) & x $\geq$ y & $<$geq$>$ \\
Greater than     & 2 & gt(x,y) & x $>$ y & $<$gt$>$ \\
Less than or equal     & 2 & le(x,y) & x $\leq$ y & $<$leq$>$ \\
Less than     & 2 & lt(x,y) & x $<$ y & $<$lt$>$ \\
\hline      
\multicolumn{2}{c}{ } \\
\multicolumn{5}{l}{Logic (operands are Booleans)} \\
\hline 
Logical not     & 1 & not(x) & $\lnot$ x & $<$not$>$ \\
Logical and     & 2 &  and(x,y) & x $\land$ y & $<$and$>$ \\
Logical or     & 2 & or(x,y) & x $\lor$ y & $<$or$>$ \\
Logical xor     & 2 & xor(x,y) & x $\oplus$ y & $<$xor$>$ \\
Logical equivalence (iff) & 2 & iff(x,y) & x $\Leftrightarrow$ y &  \\
%     & 0 & true & true \\
%     & 0 & false & false \\
\hline 
\multicolumn{2}{c}{ } \\
\multicolumn{5}{l}{Control} \\
\hline 
Alternative     & 3 & if(x,y,z) & value of y if x is true, &  \\
                &   &           & otherwise value of z     &  \\
\hline 
\end{tabular}
}
\end{center}
\caption{Operators used to build predicate expressions  \label{tab:semantics}}
\end{table}

An expression usually contains identifiers which correspond to the
formal parameters of a predicate.  To illustrate this, let us consider
the predicate that allows defining constraints involved in any
instance of the {\em queens} problem.  It corresponds to: $X \neq Y \wedge
|X-Y| \neq Z$.  We obtain using the functional representation:

{\footnotesize
\begin{verbatim}
  <predicate name="P0">
    <parameters>
      int X int Y int Z
    </parameters>
    <expression>
      <functional>
        and(ne(X,Y),ne(abs(sub(X,Y)),Z))
      </functional>
    </expression>
  </predicate>
\end{verbatim}
}

\begin{fullversion}

\subsubsection{MathML Representation}

MathML is a language dedicated to represent mathematical expressions.
We can represent predicate expressions using a subset of this language.
It is possible to insert an XML representation of the predicate expression by inserting in \xmltag{expression} an element called \xmltag{math} which contains any Boolean expression defined (in BNF notation) as follows:

{\footnotesize
\begin{verbatim}
   <integerExpression> ::= 
        "<cn>" <integer> "</cn>" 
      | "<ci>" <identifier> "</ci>" 
      | "<apply> <minus/>"  <integerExpression> "</apply>"  
      | "<apply> <abs/>"  <integerExpression> "</apply>" 
      | "<apply> <plus/>" <integerExpression> <integerExpression> "</apply>" 
      | "<apply> <minus/>" <integerExpression> <integerExpression> "</apply>" 
      | "<apply> <times/>" <integerExpression> <integerExpression> "</apply>" 
      | "<apply> <quotient/>" <integerExpression> <integerExpression> "</apply>" 
      | "<apply> <rem/>" <integerExpression> <integerExpression> "</apply>" 
      | "<apply> <power/>" <integerExpression> <integerExpression> "</apply>" 
      | "<apply> <min/>" <integerExpression> <integerExpression> "</apply>" 
      | "<apply> <max/>" <integerExpression> <integerExpression> "</apply>" 

   <booleanExpression> ::= 
        "<false/>" 
      | "<true/>"
      | "<apply> <not/>" <booleanExpression> "</apply>"
      | "<apply> <and/>" <booleanExpression> <booleanExpression> "</apply>"
      | "<apply> <or/> <booleanExpression> <booleanExpression> "</apply>"
      | "<apply> <xor/> <booleanExpression> <booleanExpression> "</apply>"
      | "<apply> <eq/>" <integerExpression> <integerExpression> "</apply>"
      | "<apply> <neq/>" <integerExpression> <integerExpression> "</apply>"
      | "<apply> <gt/>" <integerExpression> <integerExpression> "</apply>"
      | "<apply> <geq/>" <integerExpression> <integerExpression> "</apply>"
      | "<apply> <lt/>" <integerExpression> <integerExpression> "</apply>"
      | "<apply> <leq/>" <integerExpression> <integerExpression> "</apply>"
\end{verbatim}
}

For more information, see {\footnotesize \url{http://www.w3.org/Math}} or {\footnotesize \url{http://www.dessci.com/en/support/tutorials/mathml/content.htm}}.
For example, to represent $X \neq Y \wedge |X-Y| \neq Z$, we can write:

{\footnotesize
\begin{verbatim}
   <predicate name="P0">
     <parameters> int X int Y int Z </parameters>
     <expression>
       <math>
         <apply> 
            <neq/> <ci> X </ci> <ci> Y </ci>
         </apply>
         <apply> 
            <neq/>
            <apply> 
              <abs/> 
              <apply> <minus/ > <ci> X </ci> <ci> Y </ci> </apply> 
            </apply>
            <ci> Z </ci>
         </apply>
       </math>
     </expression>
   </predicate>
\end{verbatim}
}

\subsubsection{Postfix Representation}

It is possible to insert a postfix representation (not fully described in this document) of the predicate expression by inserting in \xmltag{expression} an element \xmltag{postfix}.
For example, to represent $X \neq Y \wedge |X-Y| \neq Z$, we can write:

{\footnotesize
\begin{verbatim}
   <predicate name="P0">
     <parameters>
       int X int Y int Z
     </parameters>
     <expression>
       <postfix> 
         X Y ne X Y sub abs Z ne and 
       </postfix>
     </expression>
   </predicate>
\end{verbatim}
}

\subsubsection{Infix Representation}

It is possible to insert an infix representation (not fully described in this document) of the predicate expression by inserting in \xmltag{expression} an element \xmltag{infix}.
For example, to represent $X \neq Y \wedge |X-Y| \neq Z$, we can write (using a C syntax for the Boolean expression):

{\footnotesize
\begin{verbatim}
   <predicate name="P0">
     <parameters>
       int X int Y int Z
     </parameters>
     <expression>
       <infix syntax="C"> 
         X != Y && abs(X-Y) != Z
       </infix>
     </expression>
   </predicate>
\end{verbatim}
}

\end{fullversion}

\subsection{Constraints}

The XML element called \xmltag{constraints} admits an attribute which is called \xmlattr{nbConstraints} and contains some occurrences (at least, one) of an element called \xmltag{constraint}, one for each constraint of the instance.
The attribute \xmlattr{nbConstraints} is of type integer and its value is equal to the number of occurrences of the element \xmltag{constraint}. 

Each element \xmltag{constraint} admits four attributes, called \xmlattr{name}, \xmlattr{arity}, \xmlattr{scope} and \xmlattr{reference}, and potentially contains some elements:

{\footnotesize
\begin{verbatim}
   <constraint 
      name = 'put here the name of the constraint'
      arity = 'put here the arity of the constraint' 
      scope = 'put here the scope of the constraint' 
      reference = 'put here the name of a relation, of 
                   predicate or a global constraint'>
      ...
   </constraint>
\end{verbatim}
}

The attribute \xmlattr{name} corresponds to the name of the constraint and its value must be a valid identifier.

\begin{shortversion}
\begin{competition}
The name of the $i^{th}$ constraint must  be the letter $C$ followed by the number $i-1$ (i.e. we have $C0$,  $C1$, ...).
\end{competition}
\end{shortversion}

The attribute \xmlattr{arity} is of type integer and its value is
equal to the arity of the constraint (that is to say, the number of
variables in its scope).  It must be greater than or equal to $1$.
The value of the attribute \xmlattr{scope} denotes the set of
variables involved in the constraint.  It must correspond to a list of
distinct variable names where each name corresponds to the value of
the \xmlattr{name} attribute of a \xmltag{variable} element. Variables
are separated by whitespace.

\begin{shortversion}
\begin{competition}
  To facilitate the identification of constraints which have the same
  scope, constraints in the competition instances will be sorted
  according to the lexicographic order of their normalized scope. The
  normalized scope is the scope of the constraint sorted by variable
  name (for example, \xml{V0 V1 V2} is the normalization of \xml{V2 V0
    V1}).

  This will ensure that two constraints which involve the same set
  of variables will be adjacent in the list of constraints. However,
  constraints which involve the same set of variables may have
  different scope attributes. For example, if C1 is defined on \xml{V0
    V1} and C2 is defined on \xml{V1 V0}, it is guaranteed that C1 and
  C2 will be adjacent in the list (i.e. within the element \xmltag{constraints}) but C2 will not be rewritten to have
  scope \xml{V0 V1} because this is too expensive in a number of cases.
 
  Solvers which need to identify constraints which involve the same
  sets of variables must generate the normalized scope and compare it
  to the normalized scope of the previous constraint. If they are
  equal, the constraints involve the same set of variables and they
  may be merged into a single constraint using a logical and.

  More than two consecutive constraints may involve the same set of
  variables.
\end{competition}
\end{shortversion}

There are three alternatives to represent constraints.
Indeed, it is possible to introduce:
\begin{itemize}
\item constraints in extension
\item constraints in intension
\item global constraints
\end{itemize}

\subsubsection{Constraints in extension}

The value of the attribute \xmlattr{reference} must be the name of a relation.
It means that it must correspond to the value of the \xmlattr{name} attribute of a \xmltag{relation} element.
The element \xmltag{constraint} is empty when it represents a constraint defined in extension.
For example:

{\footnotesize
\begin{verbatim}
   <constraint name="C0" scope="V0 V1" reference="R0" />
\end{verbatim}
}

\begin{shortversion}
\begin{competition}
All tuples of a relation referenced by a constraint must be valid for this constraint (i.e. any value occurring in a tuple of the relation must belong to the domain of the corresponding variable in the scope of the constraint).
\end{competition}
\end{shortversion}

\subsubsection{Constraints in intension}
 
The value of the attribute \xmlattr{reference} must be the name of a predicate.
It means that it must correspond to the value of the \xmlattr{name} attribute of a \xmltag{predicate} element.

The element \xmltag{constraint} contains an element \xmltag{parameters} when it represents a constraint defined in intension.
The element \xmltag{parameters} contains a sequence of effective parameters, each one being either an integer or a variable reference (which must occur in the scope of the constraint).
For the abridged variant, a separator is inserted between two elements.
Of course, the arity of a predicate referenced by a constraint must correspond to the number of effective parameters of this constraint.
Also, all variables occurring in the scope of a constraint referencing a predicate must occur as effective parameters of this constraint.
For example:

{\footnotesize
\begin{verbatim}
   <constraint name="C0" scope="V0 V1" reference="P0">
      <parameters>
         V0 V1 1
      </parameters>
   </constraint> 
\end{verbatim}
}

The semantics is the following.
Given a tuple built by assigning a value to each variable belonging to the scope of the constraint, the predicate expression is evaluated after replacing each occurrence of a formal parameter corresponding to an effective parameter denoting a variable with the assigned value.
The tuple is allowed iff the expression evaluates to $true$.

\begin{fullversion}
In the current version of the format, effective parameters are
restricted to be either variables in the scope of the constraint, or
integer constants. Future version of the format will remove this
limitation and allow any expression as an effective parameter.
\end{fullversion}

\subsubsection{Global constraints}

The value of the attribute \xmlattr{reference} must be the name of a
global constraint, prefixed by ``global:''.  As the character ':'
cannot occur in any valid identifier, it avoids some potential
collision with other identifiers. 
The name of global constraints is case-insensitive.
Therefore, {\tt global:allDifferent} and {\tt
  global:alldifferent} represent the same constraint.
%This is also the case for {\tt global:minimum\_weight\_alldifferent} and {\tt  global:minimumWeightAllDifferent}.

The element \xmltag{constraint} may contain an element
\xmltag{parameters} when it represents a global constraint.  If
present, the element \xmltag{parameters} contains a sequence of
parameters specific to the global constraint.  As a consequence, the
description of such parameters must be given for each global
constraint.  It is then clear that, for each global constraint, we
have to indicate its name (the one to be referenced), its parameters
(and the way they are structured in XML) and its semantics.  Below, we
provide such information for four global
constraints. % (the first one, $allDifferent$ will be considered for the next competition).

\begin{shortversion}
\begin{competition}
Only the following global constraints will be considered: $allDifferent$, $weightedSum$, $element$ and $cumulative$.
\end{competition}
\end{shortversion}

\paragraph{Constraint weightedSum (not defined in the global constraint catalog)}

\subparagraph{Semantics} $\sum_{i=1}^{r}{k_i*X_i}$ op $b$ where $r$ denotes the arity of the constraint, $k_i$ denotes an integer, $X_i$ the $i^{th}$ variable occurring in the scope of the constraint, $op$ a relational operator in $\{=,\neq,>,\geq,<,\leq\}$, and $b$ an integer.

\subparagraph{Parameters} There is a first parameter that represents a
list of $k$ dictionaries representing each product in the sum. Each
dictionary contains an integer coefficient (associated with the
\xml{coef} key) and one variable identifier (associated with the
\xml{var} key). The conventional order of keys in these dictionaries
is \xml{coef,var}. Therefore, \xml{\{/coef 2 /var X1\}} can be
represented as \xml{\{2 X1\}}.

%Space is used as a separator between pairs and between integers and variable identifiers.
There is a second parameter which is a tag denoting the
relational operator.  It corresponds to an atom that must necessarily belong to $\{$\xml{<eq/>,<ne/>,<ge/>}, \xml{<gt/>,<le/>,<lt/>}$\}$ (see Table \ref{tab:semantics}).
There is a third parameter which is an integer.

\subparagraph{Example} $V0 + 2V1 -3V2 > 12$

{\footnotesize
\begin{verbatim}
   <constraint name="C2" arity="3" scope="V0 V1 V2" reference="global:weightedSum">
     <parameters>
        [ { 1 V0 } { 2 V1 } { -3 V2 } ] 
        <gt/>
        12
     </parameters>
   </constraint> 
\end{verbatim}
}

\begin{shortversion}
\begin{competition}
All coefficients must be non null and all variable identifiers must be
different and must occur in the scope of the constraint.
\end{competition}
\end{shortversion}

The syntax of the weightedSum global constraint slightly changed from
the XCSP 2.0 format to the current format.  The first parameter of
this constraint is now a list of dictionaries (previously, it was a
list of lists).  The old syntax is deprecated.

\subparagraph{Alternatives}

This arithmetic constraint can be represented in intension (using the grammar described earlier in the paper). 
It is interesting to note that:
\begin{itemize}
\item $\sum_{i=1}^{r}{k_i*X_i} = b \Leftrightarrow \sum_{i=1}^{r}{k_i*X_i} \geq b \wedge \sum_{i=1}^{r}{k_i*X_i} \leq b$
\item $\sum_{i=1}^{r}{k_i*X_i} \neq b \Leftrightarrow \sum_{i=1}^{r}{k_i*X_i} > b \vee \sum_{i=1}^{r}{k_i*X_i} < b$
\item $\sum_{i=1}^{r}{k_i*X_i} > b \Leftrightarrow \sum_{i=1}^{r}{k_i*X_i} \geq b-1$ 
\item $\sum_{i=1}^{r}{k_i*X_i} < b \Leftrightarrow \sum_{i=1}^{r}{-k_i*X_i} > -b$ 
\end{itemize}

\subparagraph{References}

This arithmetic constraint is related to the constraint called sum\_ctr in \cite{BCR_catalog}.
Some information can also be found in \cite{RM_sum}.

\subsubsection{Global constraints from the Catalog}

The catalog of global constraints (see
\url{http://www.emn.fr/x-info/sdemasse/gccat}) describes a huge number
of global constraints. This section describes how these constraints
can be translated in the XML representation. This description is based
on the 2006-09-30 version of the catalog.

This version of the XML format supports global constraints from the catalog with parameters of type {\tt int}, {\tt dvar}, {\tt list} and {\tt collection} (in the catalog terminology). 

\begin{shortversion}
The current version of the format doesn't support constraints with {\tt sint}, {\tt svar}, {\tt mint}, {\tt mvar}, {\tt flt} or {\tt fvar} parameters (this restriction will be lifted in future versions of the format).
\end{shortversion}

Unless stated otherwise for a particular constraint, the name of the global constraint in the XML representation is
directly obtained from the name of the constraint in the catalog by
prefixing it with 'global:'. For example, the catalog defines a
constraint named {\tt cumulative}. In the XML representation, it is
named {\tt global:cumulative}. The semantics of the global constraint
is the one defined in the catalog.

Except for some particular cases, parameters of global constraints are represented according to the
following rules.

\begin{itemize}
\item[\bf atom] In the catalog, a parameter of type {\tt atom} is represented in the XML format by a tag.
Atoms representing relational operators will be denoted by elements of $\{$\xml{<eq/>,<ne/>,<ge/>}, \xml{<gt/>,<le/>,<lt/>}$\}$ (see Table \ref{tab:semantics}).

\item[\bf int] In the catalog, a parameter of type {\tt int} is an
  integer constant. It is represented in the XML format as an integer
  constant (\xml{<i>value</i>} in tagged notation or \xml{value} in
  abridged notation).

\item[\bf dvar] In the catalog, a parameter of type {\tt dvar} corresponds to a
  CSP variable. It is represented in the XML format as a variable reference (\xml{<var name="X"/>} in tagged notation or \xml{X} in
  abridged notation). A parameter of type {\tt dvar} can also correspond to an integer constant.

\item[\bf list] A list of elements in the catalog of global
  constraints is represented as a list in the XML format
  (cf. \ref{basictypes:list}).

\item[\bf collection] The catalog defines a collection as a collection
  of ordered items, each item being a set of {\tt <attribute, value>}
  pairs. In our XML representation, this is directly translated as a
  list of dictionaries. The keys in each dictionary correspond to the
  attributes in the collection.

  As an example, the {\tt cumulative} constraint is defined in the
  catalog as {\tt cumulative(TASKS; LIMIT)} where TASKS is a
  collection(origin--dvar; duration--dvar; end--dvar; height--dvar)
  and LIMIT is an int. For each task, height must be defined but only
  two attributes among origin, duration and end can be defined (since
  by definition origin-end=duration). A constraint that enforces a
  maximal height of 4 for 3 tasks  starting at origins represented as a  CSP variable and with given duration and height, 
can be represented by:

{\footnotesize
\begin{verbatim}
<constraint name="C1" arity="3" scope="X2 X5 X9" reference="global:cumulative">
  <parameters>
     [
       {/origin X2 /duration 10 /height 1} 
       {/origin X5 /duration 5 /height 2}
       {/origin X9 /duration 8 /height 3}
     ]
     4
  </parameters>
</contraint>
\end{verbatim}
}

  Note that attributes that are not required are represented as
  missing keys in the dictionaries.

  Assuming the conventional order origin, duration, end, height is defined
  for {\tt cumulative}, this constraint can also be written as

{\footnotesize
\begin{verbatim}
<constraint name="C1" arity="3" scope="X2 X5 X9" reference="global:cumulative">
  <parameters>
     [
       {X2 10 <nil/> 1} 
       {X5 5 <nil/> 2}
       {X9 8 <nil/> 3}
     ]
     4
  </parameters>
</contraint>
\end{verbatim}
}

  Here, missing attributes are represented by the \xml{<nil/>} element.

  For each global constraint, the conventional order of dictionaries
  is the order of attributes defined in the global catalog.
  
  When a global constraint has a collection parameter which contains
  only one attribute, it is represented as a list of values. For
  example, the {\tt global\_cardinality} constraint has a first parameter of
  type collection of dvar. It is represented directly a list of
  variables (\xml{[X1 X2 X3]}) instead of a list of dictionaries with
  one single key (\xml{[\{/var X1\} \{/var X2\} \{/var X3\}]} or,
  using conventional order, \xml{[\{X1\} \{X2\} \{X3\}]}).

\begin{shortversion}
  When a global constraint has a collection parameter and this
  parameter is known to contain one single item (by definition of the
  global constraint), this parameter is represented as a single
  dictionary (instead of a list containing one single dictionary).
\end{shortversion}

\end{itemize}

Arguments of a global constraint in the catalog must be translated in
XML as a sequence of elements inside the tag \xmltag{parameters} in the
order defined in the catalog. 

\begin{shortversion}
\begin{competition}
The abridged representation of global  constraints will only be given using the conventional order.
\end{competition}
\end{shortversion}

% The order in which arguments and
%attributes are given is very important as it indicates where
%information can be found.  However, in some cases (for some
%constraints), some elements (attributes) are optional. The tag
%\xml{<nil/>} is used to indicate an ommitted parameter.  This is
%illustrated below with the constraint cumulative.

\begin{shortversion}
Below, we present the description of the $3$ global constraints from the catalog that will be used for the competition (together with weightedSum).
However, note that most of the global constraints from the catalog can be directly translated in format XCSP 2.1.
Many examples (e.g. among, cycle, diffn, global\_cardinality, nvalue, etc.) are given in the document fully describing the format.
\end{shortversion}

\begin{fullversion}
Below, we present the description of some translations of global constraints from the catalog.
Note that most of the global constraints from the catalog can be automatically translated in format XCSP 2.1.
\end{fullversion}

\paragraph{Constraint allDifferent}
 
It is defined in the catalog as follows:

{\footnotesize
\begin{verbatim}
   alldifferent(VARIABLES)

   VARIABLES collection(var:dvar)
\end{verbatim}
}

\noindent Following rules given above, we may have in XML:

{\footnotesize
\begin{verbatim}
<constraint name="C1" arity="4" scope="V1 V2 V3 V4" reference="global:allDifferent">
   <parameters>
       [ V1 V2 V3 V4 ]
   </parameters>
</constraint>
\end{verbatim}
}

\begin{fullversion}
Note that it is also possible to include integer constants. For example:

{\footnotesize
\begin{verbatim}
<constraint name="C1" arity="4" scope="V1 V2 V3 V4" reference="global:allDifferent">
   <parameters>
       [ V1 V2 100 V3 V4 ]
   </parameters>
</constraint>
\end{verbatim}
}
\end{fullversion}

\begin{shortversion}
\begin{competition}
For allDifferent, parameters must only involve variables (integer constants are not accepted).
\end{competition}
\end{shortversion}

Note that the old syntax, with implicit parameters, is deprecated.
On the other hand, this constraint can be represented in intension by introducing a predicate that represents a conjunction of inequalities.
It can also be converted into a clique of binary notEqual constraints.
For more information about this constraint, see e.g. \cite{R_filtering,H_alldiff,BCR_catalog}.

\begin{fullversion}
\paragraph{Constraint among}

It is defined in the catalog as follows:

{\footnotesize
\begin{verbatim}
    among(NVAR,VARIABLES,VALUES)

    NVAR       	dvar
    VARIABLES  	collection(var:dvar)
    VALUES      collection(val:int)
\end{verbatim}
}

\noindent Following rules given above, as an illustration, we can have in XML:

{\footnotesize
\begin{verbatim}
<constraint name="C1" arity="5" scope="V0 V1 V2 V3 V4" reference="global:among">
   <parameters>
       V0
       [ V1 V2 V3 V4 ]
       [ 1 5 8 10 ]
   </parameters>
</constraint>
\end{verbatim}
}

\paragraph{Constraint atleast}
 
It is defined in the catalog as follows:

{\footnotesize
\begin{verbatim}
    atleast(N,VARIABLES,VALUE)

    N	        int
    VARIABLES	collection(var:dvar)
    VALUE     int
\end{verbatim}
}

\noindent Following rules given above, we can have in XML:

{\footnotesize
\begin{verbatim}
<constraint name="C1" arity="4" scope="V1 V2 V3 V4" reference="global:atleast">
   <parameters>
       1
       [ V1 V2 V3 V4 ]
       2 
   </parameters>
</constraint>
\end{verbatim}
}

\paragraph{Constraint atmost}
 
It is defined in the catalog as follows:

{\footnotesize
\begin{verbatim}
    atmost(N,VARIABLES,VALUE)

    N	        int
    VARIABLES	collection(var:dvar)
    VALUE    	int
\end{verbatim}
}

\noindent Following rules given above, we can have in XML:

{\footnotesize
\begin{verbatim}
<constraint name="C1" arity="4" scope="V1 V2 V3 V4" reference="global:atmost">
   <parameters>
       1
       [ V1 V2 V3 V4 ]
       2 
   </parameters>
</constraint>
\end{verbatim}
}
\end{fullversion}

\paragraph{Constraint cumulative}

It is defined in the catalog as follows:

{\footnotesize
\begin{verbatim}
   cumulative(TASKS,LIMIT)

   TASKS     	collection(origin:dvar, duration:dvar, end:dvar, height:dvar)
   LIMIT     	int
\end{verbatim}
}

\noindent Here, we have a collection of ordered items where each item corresponds to a task with $4$ attributes (origin, duration, end and height), and a limit value.
As an illustration, we can have in XML:

{\footnotesize
\begin{verbatim}
<constraint name="C1" arity="8" scope="O1 D1 E1 H1 O2 D2 E2 H2" 
            reference="global:cumulative">
   <parameters>
       [ { O1 D1 E1 H1 } { O2 D2 E2 H2 } ]
       8
   </parameters>
</constraint>
\end{verbatim}
}

As indicated in the constraint restrictions, one attribute among
$\{origin,$ $duration,end\}$ may be missing.  Assume that it is the case
for $end$, we could have:

{\footnotesize
\begin{verbatim}
<constraint name="C1" arity="6" scope="O1 D1 H1 O2 D2 H2" 
            reference="global:cumulative">
   <parameters>
       [ { O1 D1 <nil/> H1 } { O2 D2 <nil/> H2 } ]
       8
   </parameters>
</constraint>
\end{verbatim}
}

%\begin{shortversion}
%\begin{competition}
%The first parameter (the list of tasks) must only involve variables (integer constants are then not accepted).
%\end{competition}
%\end{shortversion}

\begin{fullversion}
 \paragraph{Constraint cycle}

It is defined in the catalog as follows:

{\footnotesize
\begin{verbatim}
   cycle(NCYCLE,NODES)

   NCYCLE      	dvar
   NODES       	collection(index:int, succ:dvar)
\end{verbatim}
}

\noindent Here, we have a value that denotes a number of cycles and a collection of ordered items where each item consists of $2$ attributes (index, succ).
As an illustration, we can have in XML:

{\footnotesize
\begin{verbatim}
<constraint name="C1" arity="5" scope="V0 V1 V2 V3 V4" reference="global:cycle">
   <parameters>
       V0
       [ {1 V1} {2 V2} {3 V3} {4 V4} ]
   </parameters>
</constraint>
\end{verbatim}
}

\paragraph{Constraint diffn}

It is defined in the catalog as follows:

{\footnotesize
\begin{verbatim}
   diffn(ORTHOTOPES)

   ORTHOTOPES	collection(orth:ORTHOTOPE) 
   ORTHOTOPE	collection(origin:dvar, size:dvar, end:dvar)
\end{verbatim}
}

\noindent As an illustration, we can have in XML:

{\footnotesize
\begin{verbatim}
<constraint name="C1"  arity="18" scope="V0 V1 V2 V3 V4 V5 V6 V7 V8 V9 V10 
            V11 V12 V13 V14 V15 V16 V17"  reference="global:diffn">
   <parameters>
      [ 
         [ {V0 V1 V2} {V3 V4 V5} ] 
         [ {V6 V7 V8} {V9 V10 V11} ]
         [ {V12 V13 V14} {V15 V16 V17} ]
      ]
   </parameters>
</constraint>
\end{verbatim}
}

 \paragraph{Constraint disjunctive}

It is defined in the catalog as follows:

{\footnotesize
\begin{verbatim}
   disjunctive(TASKS)

   TASKS       	collection(origin:dvar, duration:dvar)
\end{verbatim}
}

\noindent Here, we have a collection of ordered items where each item corresponds to a task with $2$ attributes (origin, duration).
As an illustration, we can have in XML:

{\footnotesize
\begin{verbatim}
<constraint name="C1" arity="6" scope="O1 D1 O2 D2 O3 D3" reference="global:disjunctive">
   <parameters>
        [ { O1 D1} {O2 D2} {O3 D3} ] 
   </parameters>
</constraint>
\end{verbatim}
}
\end{fullversion}

\paragraph{Constraint element}

It is defined in the catalog as follows:

{\footnotesize
\begin{verbatim}
   element(INDEX,TABLE,VALUE)

   INDEX        dvar
   TABLE        collection(value:dvar)
   VALUE        dvar
\end{verbatim}
}

\noindent As an illustration, we can have in XML:

{\footnotesize
\begin{verbatim}
<constraint name="C1" arity="5" scope="I X1 X2 X3 V"  reference="global:element">
   <parameters>
      I
      [ X1 X2 X3 ]
      V
   </parameters>
</constraint>
\end{verbatim}
}

%\begin{shortversion}
%\begin{competition}
%The first and second parameters must only involve variables (constant integers are then not accepted). 
%The third parameter may be a variable or an integer constant.
%\end{competition}
%\end{shortversion}

\begin{fullversion}
\paragraph{Constraint global\_cardinality}

It  is defined in the catalog as follows:

{\footnotesize
\begin{verbatim}
   global_cardinality(VARIABLES,VALUES)

   VARIABLES	collection(var:dvar)
   VALUES	collection(val:int, noccurrence:dvar)
\end{verbatim}
}

\noindent As an illustration, we can have in XML:

{\footnotesize
\begin{verbatim}
<constraint name="C1" arity="5" scope="V0 V1 V2 V3 V4"  
            reference="global:global_cardinality">
   <parameters>
       [ V0 V1 V2 ]
       [ { 1 V3 } { 2 V4 } ]
   </parameters>
</constraint>
\end{verbatim}
}

\paragraph{Constraint global\_cardinality\_with\_costs}

It is defined as follows:

{\footnotesize
\begin{verbatim}
   global_cardinality_with_costs(VARIABLES,VALUES,MATRIX,COST)

   VARIABLES	collection(var:dvar)
   VALUES	collection(val:int, noccurrence:dvar)
   MATRIX	collection(i:int, j:int, c:int)
   COST	dvar
\end{verbatim}
}

\noindent As an illustration, we can have in XML:

{\footnotesize
\begin{verbatim}
<constraint name="C1" arity="5" scope="V0 V1 V2 V3 V4" 
            reference="global:global_cardinality_with_costs">
   <parameters>
       [ V0 V1 V2 ]
       [ { 1 V3 } { 2 V4 } ]
       [ {1 1 1} {1 1 0} {1 1 3} {1 1 2} {1 1 4} {1 1 2} ]
       V5
   </parameters>
</constraint>
\end{verbatim}
}

\paragraph{Constraint minimum\_weight\_all\_different}

It is defined as follows:

{\footnotesize
\begin{verbatim}
   minimum_weight_alldifferent(VARIABLES,MATRIX,COST) 

   VARIABLES   	collection(var:dvar)
   MATRIX      	collection(i:int, j:int, c:int)
   COST        	dvar
\end{verbatim}
}

\noindent As an illustration, we can have in XML:

{\footnotesize
\begin{verbatim}
<constraint name="C1" scope="V0 V1 V2 V3"  reference="global:minimum_weight_all_different">
   <parameters>
       [ V0 V1 V2 ]
       [ {1 1 1} {1 1 0} {1 1 3} {1 1 2} {1 1 4} {1 1 2} ]
       V3
 </parameters>
</constraint>
\end{verbatim}
}

\paragraph{Constraint not\_all\_equal}

It is defined in the catalog as follows:

{\footnotesize
\begin{verbatim}
   not_all_equal(VARIABLES)

   VARIABLES	collection(var:dvar)
\end{verbatim}
}

\noindent As an illustration, we can have in XML:

{\footnotesize
\begin{verbatim}
<constraint name="C1" arity="5" scope="V0 V1 V2 V3 V4"  reference="global:not_all_equal">
   <parameters>
       [ V0 V1 V2 V3 V4 ] 
   </parameters>
</constraint>
\end{verbatim}
}

\paragraph{Constraint nvalue}

It is defined in the catalog as follows:

{\footnotesize
\begin{verbatim}
    nvalue(NVAL,VARIABLES)

    NVAL       	dvar
    VARIABLES  	collection(var:var)
\end{verbatim}
}

\noindent As an illustration, we can have in XML:

{\footnotesize
\begin{verbatim}
<constraint name="C1" arity="5" scope="V0 V1 V2 V3 V4"  reference="global:nvalue">
   <parameters>
       V0
       [ V1 V2 V3 V4 ] 
   </parameters>
</constraint>
\end{verbatim}
}

\paragraph{Constraint nvalues}

It is defined in the catalog as follows:

{\footnotesize
\begin{verbatim}
   nvalues(VARIABLES,RELOP,LIMIT)

   VARIABLES	collection(var:dvar)
   RELOP	atom
   LIMIT	dvar

   RELOP in {eq,ne,ge,gt,le,lt}
\end{verbatim}
}

\noindent As an illustration, we can have in XML:

{\footnotesize
\begin{verbatim}
<constraint name="C1" arity="5" scope="V0 V1 V2 V3 V4"  reference="global:nvalues">
   <parameters>
       [ V1 V2 V3 V4 ] 
       <gt/>
       V0
   </parameters>
</constraint>
\end{verbatim}
}
\end{fullversion}

\subsection{Illustrations}

In Figures \ref{fig:queenExtension} and \ref{fig:queenIntention}, one can see the XML representation of the {\em $4$-queens} instance.
In Figures \ref{fig:testExtension}, \ref{fig:testIntention1} and \ref{fig:testIntention2}, one can see the XML representation of a CSP instance involving $5$ variables and the $5$ following constraints:
%It illustrates non binary constraints which are formally expressed in intension by:

\begin{itemize}
\item $C0$: $X0 \neq X1$
\item $C1$: $X3 - X0 \geq 2$
\item $C2$: $X2 - X0 = 2$
\item $C3$: $X1 + 2 = |X2 - X3|$
\item $C4$: $X1 \neq X4$
\end{itemize}

Finally, in Figures  \ref{fig:magicSquare1} and \ref{fig:magicSquare2}, one can see the XML representation of the {\em $3$-magic square} instance.
The global constraints $weightedSum$ and $allDifferent$ are used.

\section{Representing QCSP instances}

This is a proposal for QCSP and QCSP$^{+}$ {\bf by M. Benedetti, A. Lallouet and J. Vautard}.
\begin{shortversion}
QCSP is not considered for the 2008 competition.
The interested reader can find the exhaustive description in the non-abridged version of this document.
\end{shortversion}

\begin{fullversion}
The Quantified Constraint Satisfaction Problem (QCSP) is an extension of CSP in which variables may be quantified universally or existentially. 
A QCSP instance corresponds to a sequence of quantified variables, called prefix, followed by a conjunction of constraints.
QCSP and its semantics were introduced  in \cite{BM_beyond}.
QCSP$^{+}$ is an extension of QCSP, introduced in \cite{BLV_qcsp} to overcome some difficulties that may occur when modelling real problems with classical QCSP.
From a logic viewpoint, an instance of QCSP+ is a formula in which (i) quantification scopes of alternate type are nested one inside the other, (ii) the quantification in each scope is \emph{restricted} by a CSP called \emph{restriction} or \emph{precondition}, and (iii) a CSP to be satisfied, called goal, is attached to the innermost scope.  An example with 4 scopes is:

%  A QCSP+ formula is a sequence of scopes such that an universal (resp. existential) scope is linked to its successor by an implication (resp. conjunction): 

\begin{eqnarray}
& & \forall X_{1} ~~ (L^{\forall}_{1}(X_{1}) \rightarrow \nonumber\\
& & ~~~~~~~ \exists Y_{1} ~~ ( L^{\exists}_{1}(X_{1},Y_{1}) \wedge \nonumber\\
& & ~~~~~~~ ~~~~~~~ \forall X_{2} ~~ (L^{\forall}_{2}(X_{1},Y_{1},
X_{2}) \rightarrow \nonumber\\
& & ~~~~~~~ ~~~~~~~ ~~~~~~~ \exists Y_{2} ~~ (L^{\exists}_{2}
(X_{1},Y_{1},X_{2},Y_{2}) \wedge ~~ G(X_{1},X_{2},Y_{1},Y_{2})) \nonumber\\
& & ~~~~~~~ ~~~~~~~ ) \nonumber\\
& & ~~~~~~~ ) \nonumber\\
& & ) \label{ex3}
\end{eqnarray}

\noindent where $X_1$, $X_2$, $Y_1$, and $Y_1$ are in general sets of variables, and each $L^{Q}_i$ is a conjunction of constraints.
A more compact and readable syntax for QCSP$^+$ employs square braces to enclose restrictions. An example with 3 scopes is as follows

$$\small \forall X_1 [L_1^\forall (X_1)]\ \exists Y_1 [L_1^\exists (X_1,Y_1)]\ \forall X_2 [L_2^\forall (X_1,Y_1,X_2)]\ \ G(X_1,Y_1,X_2)$$

\noindent which reads ``for all values of $X_1$ which satisfy the constraints $L^{\forall}_{1}(X_{1})$, there exists a value for $Y_1$ that satisfies $L^{\exists}_{1}(X_{1},Y_{1})$ and is  such that for all values for $X_2$ which satisfy $L^{\forall}_{2}(X_{1},Y_{1},X_{2})$, the goal  $G(X_{1},X_{2},Y_{1})$ is satisfied''.

%The following format is proposed for QCSP+.  
A standard QCSP can be viewed as a particular case of QCSP$^{+}$ in which all quantifications are unrestricted, i.e. all the CPSs $L^{Q}_i$ are empty.

\subsection{Presentation}

With respect to format XCSP 2.0, here is the extension to the XML element called $<$presentation$>$ in order to deal with a QCSP instance:

\begin{itemize}
  \item the attribute \xmlattr{format} must be given the value \xmlval{XCSP 2.1}.
  \item the attribute \xmlattr{type} is required and its value must be \xmlval{QCSP} or  \xmlval{QCSP+}.
\end{itemize}

All the relevant information on quantification in included in a new section called ``quantification'' (see below). 

\subsection{Quantification}

This section gives the quantification structure associated with a QCSP/QCSP$^{+}$ instance.
It essentially provides an ordered list of quantification blocks, called blocks. 
The size of this list is mandatorily declared in the attribute \xmlattr{nbBlocks}:

\begin{verbatim}
<quantification nbBlocks="b">
  put here a sequence of b blocks
</quantification>
\end{verbatim}

Notice that the order in which quantification blocks are listed inside
this XML element provides key information, as it specifies the
left-to-right order of (restricted) quantifications associated with
the QCSP/QCSP$^{+}$ instance.  Each block of a QCSP/QCSP$^{+}$
instance is represented by the following XML element:

\begin{verbatim}
<block  quantifier = 'put here the quantifier type'
        scope = 'put here a list of variable names'>

  	put here an optional list of constraints (QCSP+ only)
</block>
\end{verbatim}

\noindent where

\begin{itemize}
\item the value of the attribute \xmlattr{quantifier} must be either  ``exists'' or ``forall''. It gives the kind of quantification of all  the variables in this block;

\item the value of the attribute \emph{scope} specifies the variables which are quantified in this block. 
At least one variable must be present. 
The order in which variables are listed is not relevant;

\item for QCSP instances, the body of a block element must be empty

\item for QCSP+ instances, the body of the block element may contain
  one or more \xmltag{constraint} elements that restrict the
  quantification of the block to the sole values of the quantified
  variables which satisfy all constraints defined in the body of the
  \xmltag{block} element.

  Such constraints may only refer to variables defined in the same
  block and to variables defined in previous blocks (w.r.t. to the
  order in the enclosing \xmltag{quantification} element), but not to
  variables mentioned in later blocks;
\end{itemize}

\noindent Notice the following features of well-formed QCSP/QCSP$^+$ instances:

\begin{enumerate}
\item each variable can be mentioned in \emph{at most one} block;
\item each variable must be mentioned in \emph{at least} one block; this means the problem is \emph{closed}, i.e. no free variables are allowed\footnote{This restriction may be relaxed by future formalizations of open QCSPs.};
\end{enumerate}

\subsection{Illustrations}
Let the domain of all the variables we introduce be $\{1,2,3,4\}$.

\begin{itemize}

\item An XML encoding of the QCSP: $$\exists W,X ~ ~ \forall Y  ~~ \exists Z ~~~ W + X = Y + Z, ~~Y\neq Z$$ is given in Figure \ref{fig-qcsp}.

\item  An XML encoding of the QCSP$^{+}$: {\small $$\small\exists W,X [W+X< 8, W-X>2] ~\forall Y [W\neq Y, X\neq Y] ~\exists Z [Z<W-Y] ~ W + X = Y + Z$$} is given in Figure \ref{fig-qcsp+}.  

\end{itemize}
\end{fullversion}

\section{Representing WCSP instances}

The classical CSP framework can be extended by associating weights (or costs) with tuples \cite{BMRSVF_semiring}.
The WCSP (Weighted CSP) is a specific extension that rely on a specific valuation structure S(k) defined as follows.

\begin{definition}
S(k) is a triple $([0,\dots,k],\oplus,\geq)$ where:
\item $k \in [1,\dots,\infty]$ is either a strictly positive natural or infinity,
\item $[0,1,\dots,k]$ is the set of naturals less than or equal to $k$,
\item $\oplus$ is the sum over the valuation structure defined as: $a \oplus b = min\{k,a+b\}$,
\item $\geq$ is the standard  order among naturals. 
\end{definition}

A WCSP instance is defined by a valuation structure $S(k)$, a set of
variables (as for classical CSP instances) and a set of constraints.
A domain is associated with each variable and a cost function with
each constraint.  More precisely, for each constraint $C$ and each
tuple $t$ that can be built from the domains associated with the
variables involved in $C$, a value in $[0,1,\dots,k]$ is assigned to
$t$.  When a constraint $C$ assigns the cost $k$ to a tuple $t$, it
means that $C$ forbids $t$.  Otherwise, $t$ is permitted by $C$ with
the corresponding cost.  The cost of an instantiation of variables is the sum (using
operator $\oplus$) over all constraints involving variables
instantiated.  An instantiation is consistent if its cost is
strictly less than $k$.  The goal of the WCSP problem is to find a
full consistent assignment of variables with minimum cost.

It is rather easy to represent WCSP in XML in format XCSP 2.1.
This is described below.

\subsection{Presentation}

Here are the extensions to the XML element called \xmltag{presentation} in order to deal with a WCSP instance:

\begin{itemize}
\item the attribute \xmlattr{format} must be given the value \xmlval{XCSP 2.1}.
\item the attribute \xmlattr{type} is required and its value must be \xmlval{WCSP}.
\end{itemize}

\subsection{Relations}

For WCSP represented in extension, it is necessary to introduce ``soft'' relations.  
These relations are defined as follows:

\begin{itemize}
\item the value of the attribute \xmlattr{semantics} is set to \xmlval{soft}.
\item weighted tuples are given as described in Section \ref{sec:wtuples}
%for the abridged variant, the cost of each tuple is introduced under the form of an integer valued immediately followed by the character ':'.
\item an new attribute \xmlattr{defaultCost} is mandatory and represents  the cost of any tuple which is not explicitly listed in the relation. 
Its value belongs to $[0,\dots,k]$ where $k$ is the maximal cost of the valuation structure, defined by the attribute \xmlattr{maximalCost} of the element  \xmltag{constraints} (see below).
Note that it may be the special value 'infinity'.
\end{itemize}

%It is important to remark The first tuple of a relation must be given an explicit cost.

%\begin{fullversion}

\subsection{Functions} \label{sec:functions}

Instead of representing constraints of a WCSP in extension, it is
possible to represent them in intension by introducing cost functions.
For any tuple passed to such a function, its cost is computed and
returned.  In other words, we employ here exactly the same mechanism
as the one employed for hard constraints represented in intension. The
only difference is that a predicate returns a Boolean value whereas a cost
function must return an integer value.

If present, the XML element called \xmltag{functions} admits an
attribute which is called \xmlattr{nbFunctions} and contains some
occurrences (at least, one) of an element called \xmltag{function},
one for each function associated with at least a constraint of the
instance.  The attribute \xmlattr{nbFunctions} is of type integer and
its value is equal to the number of occurrences of the element
\xmltag{function}. The \xmltag{functions} element must be a direct
child of the \xmltag{instance} element.

%\begin{shortversion}
%\begin{competition}
%For the competition, it is mandatory to define functions before they   are referenced. 
%Therefore, the element \xmltag{functions} must be  put before the \xmltag{constraints} elements.
% Functions may not reference other functions. 
%Recursive functions  (directly of indirectly) are not allowed in the competition. 
%At  last, predicates may not use functions. These restrictions are only  meant to facilitate the introduction of functions in the solvers  implementations. 
%They will be removed in the future.
%\end{competition}
%\end{shortversion}

Each element \xmltag{function} admits two attributes, called \xmlattr{name} and \xmlattr{return}, and contains two elements, called \xmltag{parameters} and \xmltag{expression}.  
It is defined as follows:

{\footnotesize
\begin{verbatim}
   <function name = 'put here the name of the function'   return='return type'>
     <parameters>
        put here a list of formal parameters
     </parameters>
     <expression>
        Put here one (or several) representation(s) of the function expression
     </expression> 
   </function>
\end{verbatim}
}

The attribute \xmlattr{name} corresponds to the name of the function and its value must be a valid identifier. 

%\begin{shortversion}
%\begin{competition}
%For the 2008 competition, the name of the $i^{th}$ function must be  the letter $F$ followed by the number $i-1$ (i.e. we have $F0$,  $F1$, ...).
%\end{competition}
%\end{shortversion}

The attribute \xmlattr{return} indicates the return type of the
function.  In this version of the format, the only return type used is
'int'.  Then elements \xmltag{parameters} and \xmltag{expression} are
defined exactly as those defined for \xmltag{predicate} elements.  The
only difference is that the expression must be of type integer instead
of being of type Boolean.

% ??? must be able to return infinity

%\begin{shortversion}
%\begin{competition}
%For the 2008 competition, only the functional representation,  described in Section \ref{sec:predicates}, is considered.
%\end{competition}
%\end{shortversion}

%\end{fullversion}

\begin{shortversion}
\begin{competition}
Functions will not be considered.
\end{competition}
\end{shortversion}

%Finally, it is interesting to introduce a new ternary operator $if$ (used to build expressions).
%The signature of this operator is Boolean$\times$integer$\times$integer.
%Its semantics is the following: $if(x,y,z)$ is evaluated to $y$ if $x$ is true, to $z$ otherwise. 

\subsection{Constraints} \label{sec:wcsp:constraints}

For any WCSP instance, it is required to introduce an attribute
\xmlattr{maximalCost} to the element \xmltag{constraints}.  The value
of this attribute is of type integer (and must be strictly positive)
and represents the maximum cost of the WCSP framework (the $k$ value).
Remember that it corresponds to a total violation and may be equal to
'infinity' (whereas $0$ corresponds to a total satisfaction).  Also,
an optional attribute \xmlattr{initialCost} to the element
$<$constraints$>$ is introduced.  If present, the value of this
attribute is of type integer and represents a constant cost that must
be added to the sum of constraints cost.  This is the cost of the
$0$-ary constraint of the WCSP framework which is sometimes assumed
(e.g. see \cite{LS_quest}).  When not present, it is assumed to be
equal to $0$.

%For each constraint, it is required to introduce an attribute $defaultCost$ if the constraint is represented in extension by a soft relation.
%The value of this attribute is of type integer and represents the cost of any tuple not occurring in the ``soft'' referenced relation.
%When not present, it is assumed to be equal to $0$.

\begin{remark} 
When representing a WCSP instance, it is possible to refer to hard constraints (defined in extension or intension).
For such constraints, an allowed tuple has a cost of $0$ while a disallowed tuple has a cost equal to the value of the attribute \xmlattr{maximalCost}.
\end{remark}

\begin{shortversion}
\begin{competition}
  \xmlattr{maximalCost} will be a constant integer (and, so different
  from 'infinity'). \xmlattr{initialCost} will always be equal to 0
  and therefore unspecified.
\end{competition}
\end{shortversion}

\subsection{Illustration}

The representation in XCSP 2.1 of an illustrative WCSP instance is given by Figure \ref{fig:wcsp}.

%\section{Validity of Instances \label{sec:rules}}

%See Section 6 of the following document: {\small \url{http://www.cril.univ-artois.fr/~lecoutre/research/tools/tools.pdf}}.

\section*{Acknowledgments}

We would like to thank very much Nicolas Beldiceanu, Marco Beneditti, Marc van Dongen, Simon de Givry, Fred Hemery, Arnaud Lallouet, Radoslaw Szymanek and J\'er\'emie Vautard for helpful suggestions and comments.

%\bibliographystyle{plain} %alpha}
%\bibliography{globalBiblio}

%\input{examples}

\begin{figure}[p]
{\footnotesize
\begin{verbatim}
<instance>
  <presentation  name="Queens"  nbSolutions="at least 1"  format="XCSP 2.1">
     This is the 4-queens instance represnted in extension.
  </presentation>
    
  <domains nbDomains="1">
    <domain name="D0" nbValues="4">
      1..4
    </domain>
  </domains>
  
  <variables nbVariables="4">
    <variable name="V0" domain="D0"/>
    <variable name="V1" domain="D0"/>
    <variable name="V2" domain="D0"/>
    <variable name="V3" domain="D0"/>
  </variables>   
 
  <relations nbRelations="3">
    <relation name="R0" arity="2" nbTuples="10" semantics="conflicts">
       1 1|1 2|2 1|2 2|2 3|3 2|3 3|3 4|4 3|4 4
    </relation>
    <relation name="R1" arity="2" nbTuples="8" semantics="conflicts">
       1 1|1 3|2 2|2 4|3 1|3 3|4 2|4 4
    </relation>
    <relation name="R2" arity="2" nbTuples="6" semantics="conflicts">
       1 1|1 4|2 2|3 3|4 1|4 4 
    </relation>
  </relations>
    
  <constraints nbConstraints="6">
    <constraint name="C0" arity="2" scope="V0 V1" reference="R0"/>
    <constraint name="C1" arity="2" scope="V0 V2" reference="R1"/>
    <constraint name="C2" arity="2" scope="V0 V3" reference="R2"/>
    <constraint name="C3" arity="2" scope="V1 V2" reference="R0"/>
    <constraint name="C4" arity="2" scope="V1 V3" reference="R1"/>
    <constraint name="C5" arity="2" scope="V2 V3" reference="R0"/>
  </constraints>
</instance>
\end{verbatim}
}
\caption{The {\em $4$-queens} instance in extension \label{fig:queenExtension}}
\end{figure}

\begin{figure}[p]
{\footnotesize 
\begin{verbatim}
<instance>
   <presentation  name="Queens"  nbSolutions="at least 1"  format="XCSP 2.1">
     This is the 4-queens instance represented in intention.
  </presentation>
      
  <domains nbDomains="1">
    <domain name="D0" nbValues="4">
       1..4
    </domain>
  </domains>
  
  <variables nbVariables="4">
    <variable name="V0" domain="D0"/>
    <variable name="V1" domain="D0"/>
    <variable name="V2" domain="D0"/>
    <variable name="V3" domain="D0"/>
  </variables>
  
  <predicates nbPredicates="1">
    <predicate name="P0">
      <parameters> int X0 int X1 int X2  </parameters>
      <expression>
        <functional> and(ne(X0,X1),ne(abs(sub(X0,X1)),X2)) </functional>
      </expression>
    </predicate>
  </predicates>
    
  <constraints nbConstraints="6">
    <constraint name="C0" arity="2" scope="V0 V1" reference="P0">
      <parameters> V0 V1 1 </parameters>
    </constraint> 
    <constraint name="C1" arity="2" scope="V0 V2" reference="P0">
      <parameters> V0 V2 2 </parameters>
    </constraint> 
    <constraint name="C2" arity="2" scope="V0 V3" reference="P0">
      <parameters> V0 V3 2 </parameters>
    </constraint> 
    <constraint name="C3" arity="2" scope="V1 V2" reference="P0">
      <parameters> V1 V2 1 </parameters>
    </constraint> 
    <constraint name="C4" arity="2" scope="V1 V3" reference="P0">
      <parameters> V1 V3 2 </parameters>
    </constraint> 
    <constraint name="C5" arity="2" scope="V2 V3" reference="P0">
      <parameters> V2 V3 1 </parameters>
    </constraint> 
  </constraints>
</instance>
\end{verbatim}
}
\caption{The {\em 4-queens} instance in intention \label{fig:queenIntention}}
\end{figure}

\begin{figure}[p]
{\footnotesize
\begin{verbatim}
<instance>
  <presentation  name="Test"  format="XCSP 2.1">
     This is another instance represented in extension.
  </presentation>

  <domains nbDomains="3">
    <domain name="D0" nbValues="7">
       0..6
    </domain>
    <domain name="D1" nbValues="3">
      1 5 10
    </domain>
    <domain name="D2" nbValues="10">
      1..5 11..15 
    </domain>
  </domains>
  
  <variables nbVariables="5">
    <variable name="V0" domain="D0"/>
    <variable name="V1" domain="D0"/>
    <variable name="V2" domain="D1"/>
    <variable name="V3" domain="D2"/>
    <variable name="V4" domain="D0"/>
  </variables>   
 
  <relations nbRelations="4">
    <relation name="R0" arity="2" nbTuples="7" semantics="conflicts"> 
       0 0|1 1|2 2|3 3|4 4|5 5|6 6
    </relation>
    <relation name="R1" arity="2" nbTuples="25" semantics="conflicts">
       1 0|1 1|1 2|1 3|1 4|1 5|1 6|2 1|2 2|2 3|2 4|2 5|2 6|3 2|3 3|
       3 4|3 5|3 6|4 3|4 4|4 5|4 6|5 4|5 5|5 6
    </relation>
    <relation name="R2" arity="2" nbTuples="1" semantics="supports">
       5 3
    </relation>
    <relation name="R3" arity="3" nbTuples="17" semantics="supports">
       0 1 3|0 5 3|0 10 12|1 1 4|1 5 2|1 10 13|2 1 5|2 5 1|2 10 14|
       3 10 5|3 10 15|4 5 11|4 10 4|5 5 12|5 10 3|6 5 13|6 10 2 
    </relation>
  </relations>
    
  <constraints nbConstraints="5">
    <constraint name="C0" arity="2" scope="V0 V1" reference="R0"/>
    <constraint name="C1" arity="2" scope="V3 V0" reference="R1"/>
    <constraint name="C2" arity="2" scope="V2 V0" reference="R2"/>
    <constraint name="C3" arity="3" scope="V1 V2 V3" reference="R3"/>
    <constraint name="C4" arity="2" scope="V1 V4" reference="R0"/>
  </constraints>
</instance>

\end{verbatim}
}
\caption{Test Instance in extension \label{fig:testExtension}}
\end{figure}

\begin{figure}[p]
{\footnotesize
\begin{verbatim}
<instance>
   <presentation  name="Test"  format="XCSP 2.1">
     This is another instance represented in intention.
  </presentation>
      
  <domains nbDomains="3">
    <domain name="dom0" nbValues="7">
       0..6
    </domain>
    <domain name="dom1" nbValues="3">
      1 5 10
    </domain>
    <domain name="dom2" nbValues="10">
      1..5 11..15 
    </domain>
  </domains>
  
  <variables nbVariables="5">
    <variable name="V0" domain="dom0"/>
    <variable name="V1" domain="dom0"/>
    <variable name="V2" domain="dom1"/>
    <variable name="V3" domain="dom2"/>
    <variable name="V4" domain="dom0"/>
  </variables>   
 
  <predicates nbPredicates="4">
    <predicate name="P0">
      <parameters> int X0 int X1 </parameters>
      <expression>
        <functional> ne(X0,X1) </functional>
      </expression>
    </predicate>
    <predicate name="P1">
      <parameters> int X0 int X1 int X2 </parameters>
      <expression>
        <functional> ge(sub(X0,X1),X2) </functional>
      </expression>
    </predicate>
    <predicate name="P2">
      <parameters> int X0 int X1 int X2 </parameters>
      <expression>
        <functional> eq(sub(X0,X1),X2) </functional>
      </expression>
    </predicate>
    <predicate name="P3">
      <parameters> int X0 int X1 int X2 int X3 </parameters>
      <expression>
        <functional> eq(add(X0,X1),abs(sub(X3,X4))) </functional>
      </expression>
    </predicate>
  </predicates>
  ...

\end{verbatim}
}
\caption{Test Instance in intention (to be continued) \label{fig:testIntention1}}
\end{figure}

\begin{figure}[p]
{\footnotesize
\begin{verbatim}
   ...
  <constraints nbConstraints="5">
    <constraint name="C0" arity="2" scope="V0 V1" reference="P0">
      <parameters> V0 V1 </parameters>
    </constraint> 
    <constraint name="C1" arity="2" scope="V0 V3" reference="P1">
      <parameters> V3 V0 2 </parameters>
    </constraint> 
    <constraint name="C2" arity="2" scope="V0 V2" reference="P2">
      <parameters> V2 V0 2 </parameters>
    </constraint> 
    <constraint name="C3" arity="3" scope="V1 V2 V3" reference="P3">
      <parameters> V1 2 V2 V3 </parameters>
    </constraint> 
    <constraint name="C4" arity="2" scope="V1 V4" reference="P0">
      <parameters> V1 V4 </parameters>
    </constraint> 
  </constraints> 
</instance>

\end{verbatim}
}
\caption{Test Instance in intention (continued) \label{fig:testIntention2}}
\end{figure}

\begin{figure}[p]
{\footnotesize
\begin{verbatim}
<instance>
  <presentation  name="Magic Square"  format="XCSP 2.1">
     This is the magic square of order 3.
  </presentation>
    
  <domains nbDomains="1">
    <domain name="dom0" nbValues="9">
      1..9
    </domain>
  </domains>
  
  <variables nbVariables="9">
    <variable name="X0" domain="dom0"/>
    <variable name="X1" domain="dom0"/>
    <variable name="X2" domain="dom0"/>
    <variable name="X3" domain="dom0"/> 
    <variable name="X4" domain="dom0"/>
    <variable name="X5" domain="dom0"/>
    <variable name="X6" domain="dom0"/>
    <variable name="X7" domain="dom0"/> 
    <variable name="X8" domain="dom0"/>
  </variables>   
 
  <constraints nbConstraints="8">
    <constraint name="C0" arity="3" scope="X0 X1 X2" reference="global:weightedSum">
     <parameters>
        [ { 1 X0 } { 1 X1 } { 1 X2 } ]
        <eq/>
        15
     </parameters>
    </constraint>
    <constraint name="C1" arity="3" scope="X3 X4 X5" reference="global:weightedSum">
     <parameters>
        [ { 1 X3 } { 1 X4 } { 1 X5 } ]
        <eq/>
        15
     </parameters>
    </constraint>
    <constraint name="C2" arity="3" scope="X6 X7 X8" reference="global:weightedSum">
     <parameters>
        [ { 1 X6 } { 1 X7 } { 1 X8 } ]
       <eq/>
        15
     </parameters>
    </constraint>
    <constraint name="C3" arity="3" scope="X0 X3 X6" reference="global:weightedSum">
     <parameters>
        [ { 1 X0 } { 1 X3 } { 1 X6 } ]
        <eq/>
        15
     </parameters>
    </constraint>
    ...
\end{verbatim}
}
\caption{The {\em $3$-magic square} instance (to be continued) \label{fig:magicSquare1}}
\end{figure}

\begin{figure}[p]
{\footnotesize
\begin{verbatim}
     ...
    <constraint name="C4" arity="3" scope="X1 X4 X7" reference="global:weightedSum">
     <parameters>
        [ { 1 X1 } { 1 X4 } { 1 X7 } ]
        <eq/>
        15
     </parameters>
    </constraint>
    <constraint name="C5" arity="3" scope="X2 X5 X8" reference="global:weightedSum">
     <parameters>
        [ { 1 X2 } { 1 X5 } { 1 X8 } ]
        <eq/>
        15
     </parameters>
    </constraint>
    <constraint name="C6" arity="3" scope="X0 X4 X8" reference="global:weightedSum">
     <parameters>
        [ { 1 X0 } { 1 X4 } { 1 X8 } ]
        <eq/>
        15
     </parameters>
    </constraint>
    <constraint name="C7" arity="3" scope="X2 X4 X6" reference="global:weightedSum">
     <parameters>
        [ { 1 X2 } { 1 X4 } { 1 X6 } ]
        <eq/>
        15
     </parameters>
    </constraint>
    <constraint name="C8" arity="9" scope="X0 X1 X2 X3 X4 X5 X6 X7 X8" 
                reference="global:allDifferent" />
  </constraints>
</instance>
\end{verbatim}
}
\caption{The {\em $3$-magic square} instance (continued) \label{fig:magicSquare2}}
\end{figure}

\begin{fullversion}

\begin{figure}[p]
{\footnotesize
\begin{verbatim}
<instance> 
  <presentation name="ExampleQCSP" format="XCSP 2.1" type="QCSP"> 
    This is a QCSP instance. 
  </presentation> 

  <domains nbDomains="1"> 
    <domain name="D0" nbValues="4"> 
      1..4 
    </domain> 
  </domains> 

  <variables nbVariables="4" > 
    <variable name="W" domain="D0"/> 
    <variable name="X" domain="D0"/> 
    <variable name="Y" domain="D0"/> 
    <variable name="Z" domain="D0"/> 
  </variables> 

  <predicates nbPredicates="2"> 
    <predicate name="P0"> 
      <parameters> int A int B  int C int D </parameters> 
      <expression> 
        <functional> eq(add(A,B),add(C,D)) </functional> 
      </expression> 
    </predicate> 
    <predicate name="P1"> 
      <parameters> int A int B </parameters> 
      <expression> 
        <functional> ne(A,B) </functional> 
      </expression> 
    </predicate> 
    </predicates> 

  <constraints nbConstraints="2"> 
    <constraint name="C0" arity="4" scope="W X Y Z" reference="P0"> 
      <parameters> W X Y Z </parameters> 
    </constraint> 
    <constraint name="C1" arity="2" scope="Y Z" reference="P1"> 
      <parameters> Y Z </parameters> 
    </constraint> 
  </constraints> 
  
  <quantification nbBlocks="3">
    <block quantifier="exists" scope="W X" \>
    <block quantifier="forall" scope="Y" \>
    <block quantifier="exists" scope="Z" \>    
  </quantification >
</instance> 
\end{verbatim}
}
\caption{A QCSP instance}
\label{fig-qcsp}
\end{figure}

\begin{figure}[p]
{\footnotesize
\begin{verbatim}
<instance> 
  <presentation name="ExampleQCSP+" format="XCSP 2.1" type="QCSP+"> 
    This is a QCSP+ instance. 
  </presentation> 

  <domains nbDomains="1"> 
    <domain name="D0" nbValues="4"> 
      1..4 
    </domain> 
  </domains> 

  <variables nbVariables="4" > 
    <variable name="W" domain="D0"/> 
    <variable name="X" domain="D0"/> 
    <variable name="Y" domain="D0"/> 
    <variable name="Z" domain="D0"/> 
  </variables> 

  <predicates nbPredicates="4"> 
     <predicate name="myP"> 
     <parameters> int A int B int C int D </parameters> 
     <expression> 
       <functional> eq(add(A,B),add(C,D)) </functional> 
     </expression> 
    </predicate> 
    <predicate name="sum_lt"> 
      <parameters> int A int B int C</parameters> 
      <expression> 
        <functional> lt(add(A,B),C) </functional> 
      </expression> 
    </predicate> 
    <predicate name="sub_gt"> 
      <parameters> int A int B int C</parameters> 
      <expression> 
        <functional> gt(sub(A,B),C)</functional> 
      </expression> 
    </predicate> 
    <predicate name="neq"> 
      <parameters> int A int B</parameters> 
      <expression> 
        <functional> ne(A,B)</functional> 
      </expression> 
    </predicate> 
  </predicates> 
  ...
  
\end{verbatim}
}
\caption{A QCSP$^{+}$ instance (to be continued)}
\label{fig-qcsp+}
\end{figure}

\begin{figure}[p]
{\footnotesize
\begin{verbatim}
  ...
  <constraints nbConstraints="1"> 
    <constraint name="goal" arity="4" scope="W X Y Z" reference="myP"> 
      <parameters> W X Y Z </parameters> 
    </constraint> 
  </constraints> 
  
  <quantification nbBlocks="3">
    <block quantifier="exists" scope="W X"> 
      <constraint name="restr1_c1" arity="2" scope="W X" reference="sum_lt"> 
        <parameters> W X 8 </parameters> 
      </constraint> 
      <constraint name="restr1_c2" arity="2" scope="W X" reference="sub_gt"> 
        <parameters> W X 2 </parameters> 
      </constraint> 
    </block>

    <block quantifier="universal" scope="Y">
      <constraint name="restr2_c1" arity="2" scope="W Y" reference="neq"> 
        <parameters> W Y </parameters> 
      </constraint> 
      <constraint name="restr2_c2" arity="2" scope="X Y" reference="neq"> 
        <parameters> X Y </parameters> 
      </constraint> 
    </block>

    <block quantifier="existential" scope="Z"> 
      <constraint name="restr3_c1" arity="3" scope="W Y Z" reference="sub_gt"> 
        <parameters> W Y Z </parameters> 
      </constraint>     
    </block> 
  </quantification >
</instance> 
\end{verbatim}
}
\caption{A QCSP$^{+}$ instance (continued)}
\label{fig-qcsp+partB}
\end{figure}

\end{fullversion}

\begin{figure}[p]
{\footnotesize
\begin{verbatim}
<instance>
  <presentation  name="ExampleWCSP"  format="XCSP 2.1"  type="WCSP">
     This is a WCSP instance.
  </presentation>

  <domains nbDomains="1">
    <domain name="D0" nbValues="3">0..2</domain>
  </domains>

  <variables nbVariables="4">
    <variable name="V0" domain="D0"/>
    <variable name="V1" domain="D0"/>
    <variable name="V2" domain="D0"/>
    <variable name="V3" domain="D0"/>
  </variables>

  <relations nbRelations="6">
    <relation name="R0" arity="2" nbTuples="10" semantics="soft" defaultCost="0">
      5:0 0|0 1|1 0|1 1|1 2|2 1|2 2|2 3|3 2|3 3
    </relation>
    <relation name="R1" arity="1" nbTuples="2" semantics="soft" defaultCost="0">
      1:1|3
    </relation>
    <relation name="R2" arity="1" nbTuples="2" semantics="soft" defaultCost="0">
      1:1|2
    </relation>
    <relation name="R3" arity="1" nbTuples="2" semantics="soft" defaultCost="0">
      1:0|2
    </relation>
  </relations>

   <functions nbFunctions="2">
     <function name="F0" return="int">
       <parameters> int X int Y </parameters>
       <expression>
         <functional> if(eq(X,Y),0,5) </functional>
       </expression>
     </function>
     <function name="F1" return="int">
       <parameters> int X int Y int Z  </parameters>
       <expression>
         <functional> if(gt(mul(add(X,Y),Z),5),0,2) </functional>
       </expression>
     </function> 
   </functions>

  <constraints nbConstraints="7" initialCost="0" maximalCost="5">
    <constraint name="C0" arity="2" scope="V0 V1" reference="R0" />
    <constraint name="C1" arity="2" scope="V0 V2" reference="F0 />
    <constraint name="C2" arity="3" scope="V1 V2 V3" reference="F1" />
    <constraint name="C3" arity="1" scope="V0" reference="R1" />
    <constraint name="C4" arity="1" scope="V1" reference="R2" />
    <constraint name="C5" arity="1" scope="V2" reference="R3" />
  </constraints>
</instance>
\end{verbatim}
}
\caption{A WCSP instance \label{fig:wcsp}}
\end{figure}

%This paper has been supported by the CNRS, the ``programme Cocoa de la R\'egion Nord/Pas-de-Calais'' and by the ``IUT de Lens''.

%\appendix

%\bibliographystyle{plain} %alpha}
%\bibliography{../globalBiblio}

\end{document}